\documentclass{article}

\usepackage{microtype}
\usepackage{graphicx}
\usepackage{tabularx}
\usepackage{arydshln}
\usepackage{multirow}
\usepackage{nicefrac}
\usepackage{enumitem}
\usepackage{subfigure}
\usepackage[algo2e]{algorithm2e}
\usepackage{setspace}
\usepackage{tikz}
\usepackage{xcolor}
\usepackage{soul}
\usepackage[bottom]{footmisc}
\RestyleAlgo{ruled}
\usepackage{booktabs}
\usepackage{amsfonts,amsmath,amssymb,amsthm}
\newcolumntype{C}{>{\centering\arraybackslash}X}
\newcolumntype{R}{>{\raggedleft\arraybackslash}X}
\usepackage{stfloats}
\renewcommand{\Re}{\operatorname{Re}}

\usepackage{hyperref}

\usepackage[accepted]{icml2024}

\icmltitlerunning{Balanced Resonate-and-Fire Neurons}

\begin{document}

\twocolumn[
\icmltitle{Balanced Resonate-and-Fire Neurons}

\begin{icmlauthorlist}
\icmlauthor{Saya Higuchi}{ul}
\icmlauthor{Sebastian Kairat}{ul}
\icmlauthor{Sander M. Bohté}{cwi}
\icmlauthor{Sebastian Otte}{ul}
\end{icmlauthorlist}

\icmlaffiliation{ul}{Adaptive AI Lab, Institute of Robotics and Cognitive Systems\\University of Lübeck, Germany}
\icmlaffiliation{cwi}{Machine Learning Group, Centrum Wiskunde \& Informatica (CWI)\\Amsterdam, The Netherlands}

\icmlcorrespondingauthor{Saya Higuchi}{sa.higuchi@uni-luebeck.de}
\icmlcorrespondingauthor{Sebastian Otte}{sebastian.otte@uni-luebeck.de}

\icmlkeywords{Spiking Neural Networks, Resonate-and-Fire, Recurrent Neural Networks}

\vskip 0.3in
]

\printAffiliationsAndNotice{}

\begin{abstract}
The resonate-and-fire (RF) neuron, introduced over two decades ago, is a simple, efficient, yet biologically plausible spiking neuron model, which can extract frequency patterns within the time domain due to its resonating membrane dynamics. However, previous RF formulations suffer from intrinsic shortcomings that limit effective learning and prevent exploiting the principled advantage of RF neurons. Here, we introduce the balanced RF (BRF) neuron, which alleviates some of the intrinsic limitations of vanilla RF neurons and demonstrates its effectiveness within recurrent spiking neural networks (RSNNs) on various sequence learning tasks. We show that networks of BRF neurons achieve overall higher task performance, produce only a fraction of the spikes, and require significantly fewer parameters as compared to modern RSNNs. Moreover, BRF-RSNN consistently provide much faster and more stable training convergence, even when bridging many hundreds of time steps during backpropagation through time (BPTT). These results underscore that our BRF-RSNN is a strong candidate for future large-scale RSNN architectures, further lines of research in SNN methodology, and more efficient hardware implementations.

\end{abstract}

\section{Introduction}
\label{intro}

Artificial neural networks (ANNs) have become the main method for solving machine learning problems in recent years \citep{Goodfellow-et-al-2016,wu2018development,abiodun2019comprehensive}.
However, they require massive computation and energy, making them inefficient for large-scale real-world applications---particularly in the realm of edge computing---
due to the deep non-linear continuous estimators to represent features of the data \cite{pfeiffer2018deep}.

Spiking neural networks (SNNs) circumvent this drawback by processing information through the precise timing of action potentials, or {\em spikes}. SNNs can potentially be more efficient compared to ANNs due to their event-driven property, for which computation is only required when spikes are propagated in the system \citep{paugam2012computing}. Furthermore, SNNs are self-recurrent, as the neurons have a dynamic internal state that modulates their activity over time, and are more biologically realistic than conventional ANNs.
Some examples of spiking neurons with ascending biological plausibility include the leaky integrate-and-fire (LIF) neuron, the adaptive leaky integrate-and-fire (ALIF) neuron \cite{bellec2018long}, the Izhikevich neuron \citep{izhikevich2003simple}, and the Hodgkin-Huxley (HH) neuron \citep{hodgkin1952quantitative}. While the HH model is too computationally expensive for practical application, it exhibits rich and complex membrane dynamics that simpler models lack. 

One mode of dynamics gaining interest in ML applications in particular is the resonating behavior seen in oscillatory neurons. Subthreshold oscillations of the membrane potential have been observed in various mammalian nervous systems, including the neurons in the frontal cortex \citep{llinas1991vitro}, in the thalamus \citep{pedroarena1997dendritic}, and in layer II of the medial entorhinal cortex \citep{alonso1989subthreshold}, which is crucial in spatial information processing \citep{doeller2010evidence}. The resonate-and-fire (RF) neuron proposed by \citet{izhikevich2001resonate} models such damped or sustained subthreshold oscillations of biological neurons: the RF neuron fires when the frequency of the incoming spikes matches that of the neuron's damped oscillation. Incoming signals with higher frequency leads to more firing for a LIF neuron but less firing for an RF neuron with slow oscillation. As the RF neuron is similarly computationally efficient as the LIF neuron, it is a spiking neuron model potentially suitable for large-scale SNNs \citep{izhikevich2001resonate}. 

Previous works showed RF neurons implemented in the Intel Loihi 2 \citep{davies2018loihi}, a neuromorphic processor, can be used to compute the short-time Fourier transform (STFT) of signals \citep{frady2022efficient,shrestha2023efficient}. Moreover, the RF neurons successfully converted raw data signals to spike trains \citep{shaaban2024resonate}, and further input to SNNs with LIF neurons for detection and classification tasks \citep{hille2022resonate, lehmann2023direct}. 
RF neurons have also been implemented in the framework of (feedforward) SNNs as harmonic oscillators for image classification tasks \citep{alkhamissi2021deep}, as well as in optical flow estimation and audio classification tasks \citep{frady2022efficient}. Still, the performance of the RF models did not significantly exceed that of deep LIF SNNs \citep{frady2022efficient} or LSTM cells \citep{alkhamissi2021deep},  and these studies did not compare with state-of-art spiking neuron models, such as ALIF. 
Considering the strengths of the RF neuron, these results suggest that the full potential of the RF neuron has yet to be explored.

Recent methodological advances in training recurrent SNNs (RSNNs) have demonstrated their potential for effective time series learning, particularly in combination with BPTT \citep{bellec2018long,yin2021accurate}. Nonetheless, they do face a ``convergence dilemma''---it requires usually up to many hundreds of epochs for RSNNs to converge properly \citep{yin2021accurate,fang2021,zhang2023long}.

Here we propose a novel spiking neuron model---the balanced RF neuron---which overcomes both, the intrinsic limitations of vanilla RF neurons as well as shortcomings that arise during training of ALIF-RSNNs. As a result, our RF variants achieve not only comparable and higher task performance than the ALIF networks, but also remarkably fast and stable convergence, reaching 95\% of the mean final accuracy within the first five epochs with up to seven times less spikes than ALIF networks.

\section{Resonate-and-Fire Neurons}
The oscillatory behavior of the membrane potential in an RF neuron is formulated with two linear differential equations: 
\begin{align}
    \dot{x} &= b\, x - \omega \, y + I \label{eq:real}\\ 
    \dot{y} &= \omega \, x +b \, y \label{eq:imag}
\end{align}
which is expressed as a single complex equation: 
\begin{equation}\label{equation:2.1}
    \dot{u} = (b + i \, \omega) \, u  + I
\end{equation}
with $u = (x + i \, y) \in \mathbb{C}$ and $I$ the injected current \cite{izhikevich2001resonate}. $\omega > 0$ is the angular frequency of the neuron, which describes how many radians the neuron progresses per second. $b < 0$ is the dampening factor that exponentially decays the oscillation. The smaller $b$ is, the faster the oscillation dampens to the resting state.

\begin{figure}[t]
    \centering
   \begin{tikzpicture}[]
    \node[anchor=south west] (image) at (0,0) {\includegraphics[width=\linewidth]
    {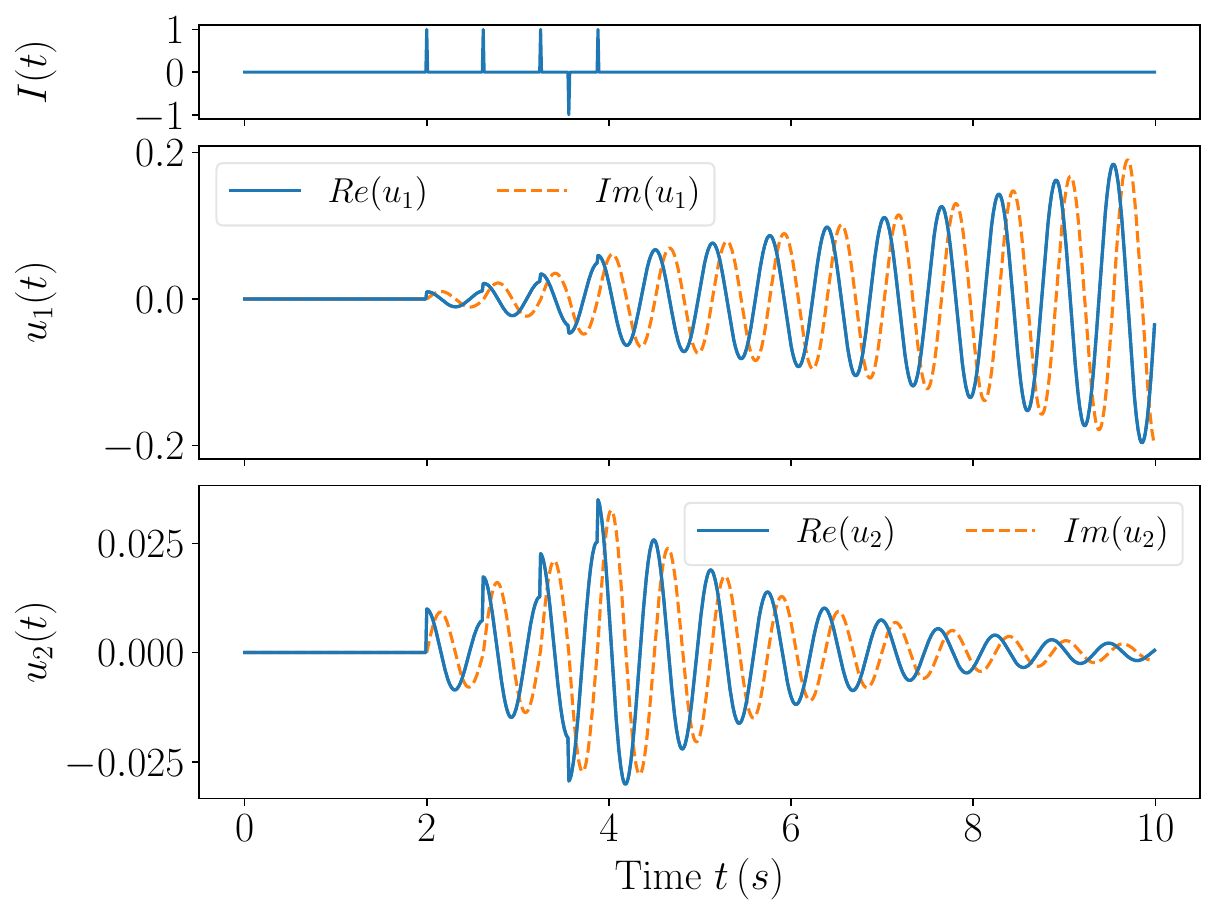}};
    \node[font=\scriptsize\linespread{0.9}\selectfont,align=left] at (2.7, 3.55) {unstable \\ ($\omega{=}10$, $b{=}-0.3$)};
    \node[font=\scriptsize\linespread{0.9}\selectfont,align=left] at (2.6, 1.21) {stable \\ ($\omega{=}10$, $b{=}-1$)};
  \end{tikzpicture}
    \caption{\label{figure:izhikevich_divconv}Membrane dynamics of two RF neurons. The four excitatory input spikes are in phase with the neuron's angular frequency $\omega = 10$. The inhibitory spike at half-phase enhances the sensitivity of the neuron to forthcoming excitatory input. Depending on its parameterization, the neuron exhibits divergence behavior (top) with $b = -0.3$ and $\delta = 0.01$ or convergence behavior (bottom) with $b = -1$. $I(t) \in \mathbb{R}$ and $u_1(t),u_2(t) \in \mathbb{C}$ refer to the injected current and the membrane potential of the neuron, respectively.}
\end{figure}

\paragraph{Izhikevich RF Neuron. }
We applied the Euler method \citep{atkinson1989introduction} to \autoref{equation:2.1} with step size (time scale) $\delta$:
\begin{equation}\label{eq:mem_update}
     u(t) = u(t-\delta) + \delta \, ((b + i\, \omega) \, u(t-\delta ) + I(t))
\end{equation}
which is used to simulate the RF neuron in \autoref{figure:izhikevich_divconv}. It shows the oscillatory behavior when similar frequency-timed input is injected into an RF neuron with an angular frequency of 10 $rad/s$ and a dampening factor of -1. 

\paragraph{Harmonic RF Neuron. }
In the harmonic RF (HRF) neuron, the membrane potential changes based on the dynamics of dampened harmonic oscillation \cite{alkhamissi2021deep}. Instead of using a complex representation, the membrane potential is split into two states as follows:
\begin{align}
\dot u &=  -2b \,u - \omega^2\,v + I \\
\dot v &= u
\end{align}
By applying Euler integration we get the equations for discrete time steps:
\begin{align} \label{Eq:hrfupdates}
    \begin{split}
    u(t) &= u(t-\delta) + \delta(-2 b \, u(t-\delta) \\
    & \qquad -
    \omega^2 v(t-\delta) + I(t))
    \end{split}
    \\
    v(t) &=v(t-\delta) + \delta u(t- \delta).\label{eq:hrfupdate2}
\end{align}
Here, $b > 0 $ is the dampening factor, and $\omega > 0$ is the angular frequency.

\section{Balanced RF Models}

Initial exploration of the RF neurons in RSNN with a random subset of 32 samples from the sequential MNIST dataset \citep{deng2012mnist} showed excessive spiking, divergent behavior, and the traditional hard and soft reset (\autoref{eq:soft-hard}) hindering immediate resonance behavior.

Divergence is due to the approximation of the dynamic system in discrete steps and is dependent on the combination of $\omega$, $b$, and the discrete-time scale $\delta$, as shown in \autoref{figure:izhikevich_divconv} (top). The membrane potential diverges and continuously produces spikes unrelated to the input signal, causing noise and artificial signals that disrupt the original frequencies, which may have hindered effective learning of the model.

\paragraph{Balanced Izhikevich RF Neuron.}
Considering the intrinsic limitations of the basic RF neuron, we introduce the balanced RF {\em (BRF)} neuron with a variation in the threshold, reset mechanism, and divergence boundary. To reduce continuous spiking of the neuron and induce spiking sparsity, a \textbf{refractory period}, referred to as $q(t)$, is implemented in the threshold,
increasing it after a neuron fires: 
\begin{align}
\vartheta(t) &= \vartheta_c + q(t)\\
z(t) &= \Theta \, (\Re(u(t)) - \vartheta(t))
\end{align}

with $\vartheta_c$ the constant threshold and $z(t)$ the output spiking. The real part of the  
\autoref{equation:2.1} was considered for the threshold mechanism, as it induces an immediate response. The refractory period decays exponentially with time: 
\begin{equation}
q(t) = \gamma q(t-\delta ) + z(t - \delta )
\end{equation}
The default refractory period constant is $\gamma = 0.9$. 

Another limitation of the basic RF model was the traditional reset mechanism, which reduces the amplitude but disrupts the oscillation. Hence, we propose the \textbf{smooth reset} as an alternative that temporarily increases the dampening of the amplitude to decay faster after the neuron fires by means of integrating the refractory period into the dampening term:
\begin{equation}
b(t) = b_c - q(t)
\end{equation}
with $b_c$ the constant dampening factor. 

\begin{figure}[t]
    \centering
    \includegraphics[width=\linewidth]{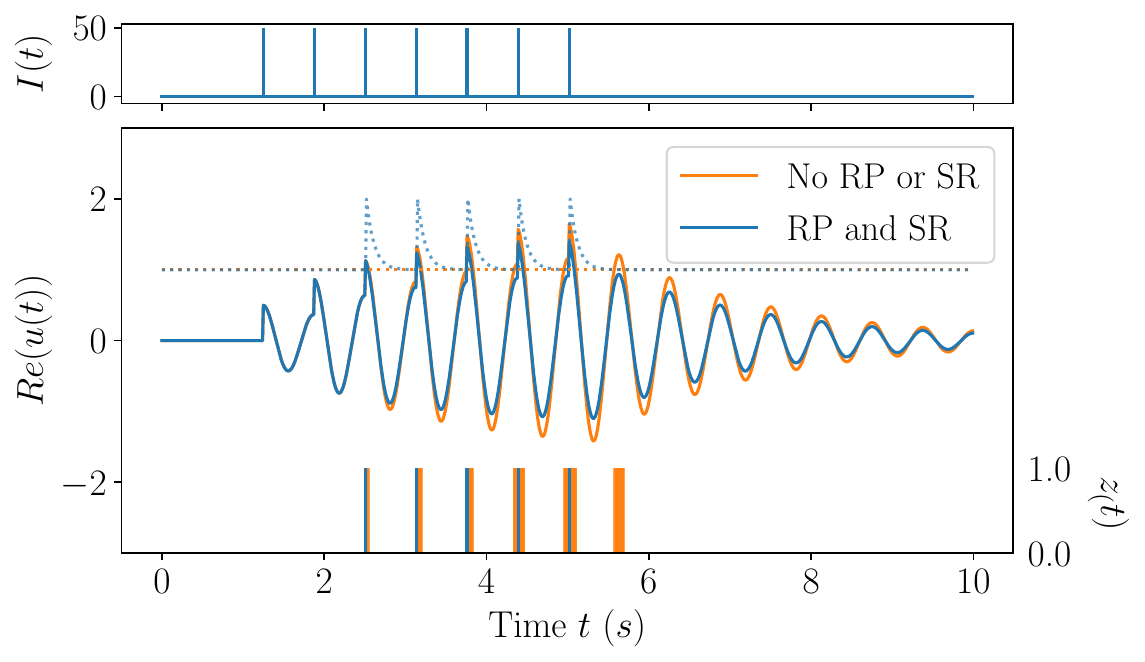}
    
    \caption{Membrane potential $u(t)$ and spiking response $z(t)$ of an RF neuron without refractory period (RP) or smooth reset (SR) (orange) compared to both (blue) for the given input signal $I(t)$.}
    \label{figure:thesis_RF}
\end{figure}

The influence of implementing the refractory period and the smooth reset can be seen in \autoref{figure:thesis_RF}. The single RF neuron was simulated with $\omega = 10, \, b = -1$, and the input signal was frequency-timed. The combination of refractory period and smooth reset significantly reduced the number of spikes, while the output spikes effectively reflected the period of the angular frequency of the neuron. The effect of implementing both are highlighted in \autoref{app:rf_sp_sop_plot} in \autoref{sec:sop_rf_smr}.

To alleviate the divergence problem, we propose to restrict the parameter space to a subspace below the \textbf{divergence boundary}---an analytically derived relation between $\delta$, $b_c$, and $\omega$ that ensures convergence.
For an RF neuron to converge or show sustained oscillation, the magnitude of the membrane potential decreases or stays constant with time given an incoming signal:
\begin{equation}
|u(t)| \leq |u(t - \delta )|
\end{equation}
After spike onset the magnitude converges---but does not precisely become 0, $|u(t-\delta )| \neq 0$, thus both side is divisible by $|u(t-\delta )|$
\begin{equation}
\frac{|u(t)|}{|u(t - \delta )|} \leq 1
\end{equation}
Furthermore, the explicit form of the subthreshold oscillatory behavior of the membrane potential after an incoming spike is derived in  \autoref{app:explicit_subthreshold} and given as: 
\begin{equation}
u(t) = \delta \, (1 + \delta (b_c + i \, \omega))^{\frac{t}{\delta}- 1}
\end{equation}
The refractory period and the reset are not considered in the derivation, as post-spiking behavior is irrelevant for modeling subthreshold behavior. The explicit form is further implemented into the inequality above and simplified: 
\begin{align*}
&|1 +  \delta (b_c + i \, \omega)| \leq 1\\
\iff &\sqrt{(1 + \delta b_c)^2 + (\delta\omega)^2} \leq 1
\end{align*}
Since the radicand is positive, both sides of the inequality can be squared:
\begin{equation}
\Longrightarrow (1 + \delta b_c)^2 + (\delta\omega)^2 -1 \leq 0
\end{equation}
The condition for the neuron to show damped oscillation is found by solving the quadratic inequality for $b_c$, where $b_c$ is in the following range of the solutions given a constant $\omega$: 
\begin{equation}
\frac{-1 - \sqrt{1 - (\delta \, \omega)^2}}{\delta} < b_c < \frac{-1 + \sqrt{1 - (\delta \, \omega)^2}}{\delta}
\end{equation}
Furthermore, the neuron is a sustained oscillator if: 
\begin{equation}
b_c = \frac{-1 \pm \sqrt{1 - (\delta \, \omega)^2}}{\delta}
\end{equation}
which leads us to another condition of the neuron, since $b_c \in \mathbb{R}_{<0}$ and $\omega \in \mathbb{R}_{>0}$:
\begin{equation} \label{eq:omega_bound}
\sqrt{1 - (\delta \, \omega)^2} > 0 \Rightarrow \omega \leq \frac{1}{\delta}
\end{equation}
With a default $\delta = 0.01$, the highest frequency that the neuron can resonate at is the frequency corresponding to an angular frequency of $100\,rad/s$.
We define this angular frequency as the upper boundary $\omega_{ub}$. The upper bound of $b_c$ that leads to sustained oscillation is implemented in this paper as $p(\omega)$, also considered the divergence boundary:
\begin{equation}\label{equation:sust}
    p(\omega) = \frac{-1 + \sqrt{1 - (\delta \, \omega)^2}}{\delta}
\end{equation}
combined with a trainable b-offset $b' > 0$ to ensure flexibility and convergence: $b_c = p(\omega) - b'$, which is constant throughout one sequence length. \autoref{fig:delta_div_bd} shows exemplary divergence boundaries for various $\delta$ values.
By combining the derived dampening factor with the smooth reset, we present the final equation of $b(t)$ applied for optimization:
\begin{equation}
b(t) = p(\omega) - b' - q(t)
\end{equation}

Implementing the refractory period, smooth reset, and the divergence boundary leads to efficient learning and sparse spiking for all four datasets. 

\paragraph{Balanced Harmonic RF Neuron. }

Similarly, we propose the balanced HRF (BHRF) neuron with the refractory period, smooth reset, and a tailored
divergence boundary for $b$ depending on $\omega$: 
\begin{equation}
 p(\omega) = \frac{\omega^2}{200}
\end{equation}
For $\delta = 0.01$ and $b_c = p(\omega)$, the BHRF neuron shows sustained oscillation.

\begin{figure}[t]
    \centering
    \includegraphics[width=\linewidth]{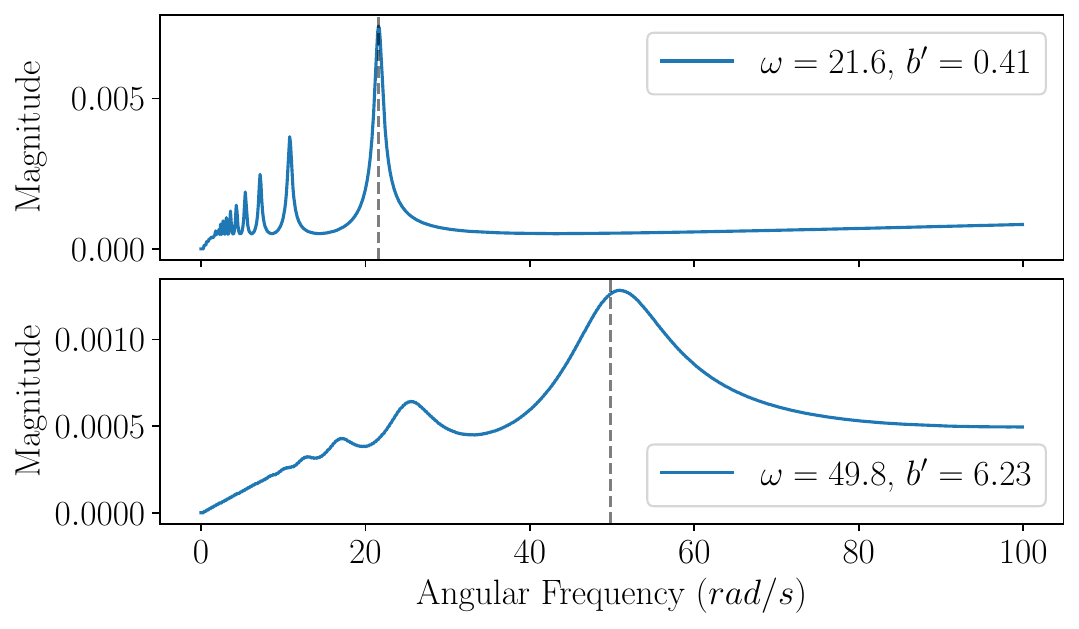}
    \caption{RF neuron frequency response plots for exemplary omega $\omega$ and b-offset $b'$ combinations with $\delta = 0.001$.} 
    \label{figure:freq_resp}
\end{figure}

\paragraph{Frequency Response of the BRF Neuron. }
The frequency response plots shown in \autoref{figure:freq_resp} assess how well the RF neuron responds to specific frequencies. A peak on a frequency response plot indicates a high sensitivity of the neuron to the respective input signal frequency (details in \autoref{app:freq_method}). The figure shows alignment of the response peak and $\omega$ of the RF neuron, demonstrating the resonance property to also be present in the discrete case. The slight offset of the response peak and $\omega$ of the RF neuron is an artifact of the numerical integration of the membrane dynamics' differential equation and disappears with decreasing the temporal step size $\delta$.

The RF neuron can also be considered a narrow band pass filter, for which only a specific range of frequencies are filtered depending on its angular frequency $\omega$. Combined with the discrete time scale $\delta$ and dampening parameter $b'$, they determine how narrow and sensitive the band pass filter is. It should be noted that RF neurons are naturally sensitive to the lower order subharmonics of their angular frequency, that is (with decaying sensitivity) $\frac{1}{2}\omega, \frac{1}{3}\omega, \frac{1}{4}\omega$ and so forth.

We focused on the BRF neuron and the BHRF remained preliminary in its exploration, as the BRF responses were more sensitive to the input frequencies than BHRF responses shown in \autoref{app:freq_plot_bhrf}.

\section{Network Implementation}

We implement the RF, BRF, and BHRF neurons within RSNNs and applied them to simulations with several benchmark datasets\footnote{Source code avaliable at \url{https://github.com/AdaptiveAILab/brf-neurons}}. The essential formulations of the BRF and the BHRF neuron are summarized in \autoref{alg:brf} and \autoref{alg:bhrf}, respectively. Note that for the algorithmic formulation we change the notation to time-discrete tensor operations and superscript the time index, whereas the transition from $t$ to $t+1$ is considered as a time delay of $\delta$.

\begin{algorithm}[b!]
\setstretch{1.2}
\setlength{\abovedisplayskip}{0pt}
\setlength{\belowdisplayskip}{0pt}
\caption{BRF Forward Pass}
\label{alg:brf}
$\mathbf{b}^{t} = p(\boldsymbol{\omega}) - \mathbf{b}' - \mathbf{q}^{t-1}$\\
$\mathbf{u}^t = \mathbf{u}^{t - 1} + \delta ((\mathbf{b}^t + i \boldsymbol{\omega}) \, \mathbf{u}^{t-1} + \mathbf{x}^{t})$\\
$\boldsymbol{\vartheta}^{t} = \vartheta_c + \mathbf{q}^{t-1}$\\
$\textbf{z}^t = \Theta (\Re(\textbf{u}^t) - \boldsymbol{\vartheta}^t)\\$
$\mathbf{q}^t = \gamma \mathbf{q}^{t-1} + \mathbf{z}^t$\\
\,\\
\small
$\vartheta_c = 1$, $\gamma=0.9$, and $p(\boldsymbol{\omega}) = \frac{-1 + \sqrt{1 - (\delta \, \boldsymbol{\omega})^2}}{\delta}$\\
($\Re$, $\Theta$, $p$ are applied component-wise.)
\end{algorithm}
\begin{algorithm}[b!]
\setstretch{1.2}
\setlength{\abovedisplayskip}{0pt}
\setlength{\belowdisplayskip}{0pt}
\caption{BHRF Forward Pass}
\label{alg:bhrf}
$\mathbf{b}^{t} = p(\boldsymbol{\omega}) - \mathbf{b}' - \mathbf{q}^{t-1}$\\
$\mathbf{u}^{t} = \mathbf{u}^{t-1} + \delta(-2 \mathbf{b}^{t}\mathbf{u}^{t-1} - \boldsymbol{\omega}^2 \mathbf{v}^{t-1} + \mathbf{x}^{t})$\\
$\mathbf{v}^{t} = \mathbf{v}^{t-1} + \delta \mathbf{u}^{t-1}$\\
$\boldsymbol{\vartheta}^{t} = \vartheta_c + \mathbf{q}^{t-1}$\\
$\textbf{z}^t = \Theta (\textbf{u}^t - \boldsymbol{\vartheta}^t)\\$
$\mathbf{q}^t = \gamma \mathbf{q}^{t-1} + \mathbf{z}^t$\\
\,\\
\small
$\vartheta_c = 1$, $\gamma=0.9$, and $p(\boldsymbol{\omega}) = \frac{\boldsymbol{\omega}^2}{200}$\\
\small($\Re$, $\Theta$, $p$ are applied component-wise.)
\end{algorithm}

We trained the networks with BPTT with the double-Gaussian function (\autoref{eq:multi-gaussian}, \citeauthor{yin2021accurate}, \citeyear{yin2021accurate}) as the surrogate gradient \cite{bellec2018long,neftci2019surrogate}. Further details of the networks are described in \autoref{app:network}.

\begin{figure*}[t]
    \centering
    \includegraphics[width=0.98\linewidth]{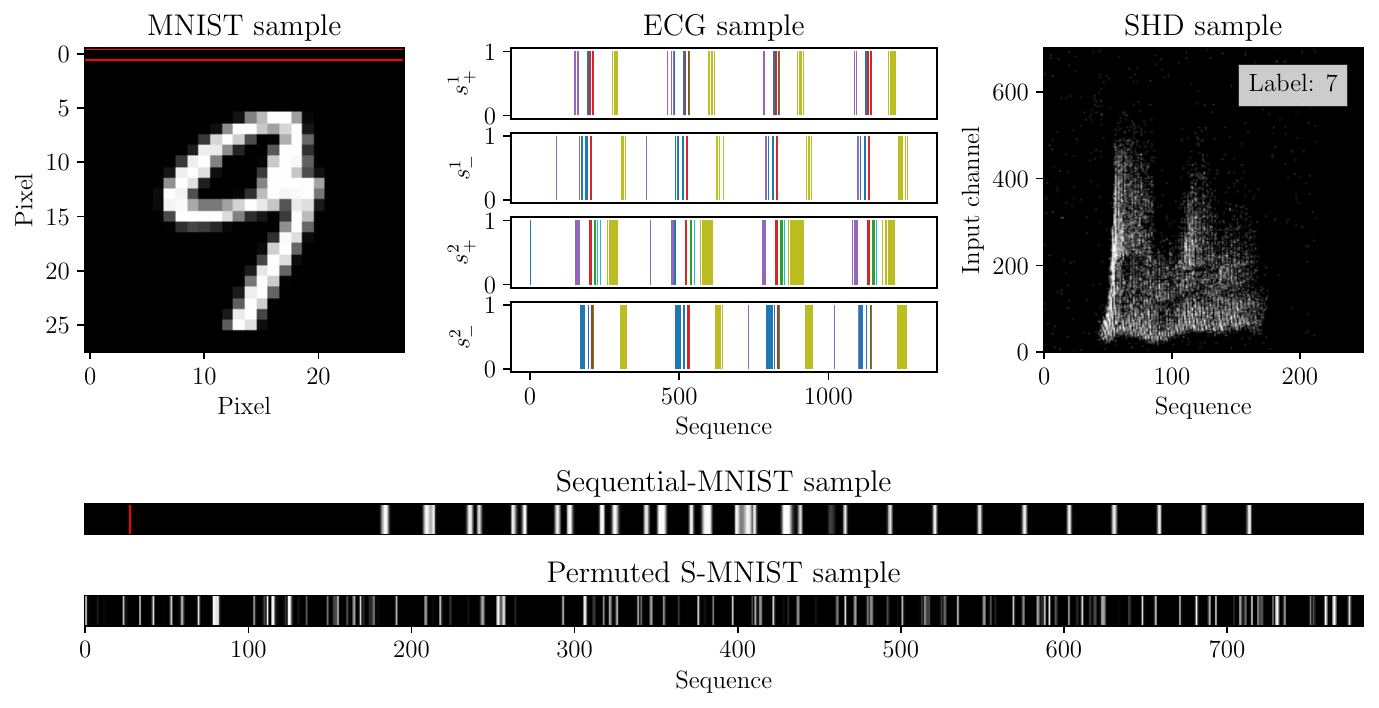}
    \caption{\label{figure:dataset}Overview of the datasets. Examplary MNIST image and its corresponding sequential and permuted representations. Common pixel row outlined in red on MNIST and S-MNIST sample. ECG sample after level cross encoding. SHD sample after preprocessing.}
\end{figure*}

\section{Experiments}
The MNIST dataset consists of grayscale 28$\times$28 pixel hand-written digit images for classification. The sequential-MNIST (\textbf{S-MNIST}), which converts the image to a sequence of 1 $\times$ 784, is a prominent benchmark dataset that enable comparison between sequential models with 54,000 images for training, 6,000 for validation, and 10,000 for testing. In the permuted S-MNIST variant (\textbf{PS-MNIST}), the pixel positions were first permuted randomly  and then input to the network sequentially.

Electrocardiogram (\textbf{ECG}) recordings of the heart are represented in voltage over time and consist of cyclic activity with six characteristic waveforms: P, PQ, QR, RS, ST, and TP. The QT database consisted of ECG recordings with per-time step labels classified by experts in the field \cite{qtdb}. There were 105 recordings, each recorded for 15 minutes with two electrodes. We left out 24 signals, as they were sudden death recordings without annotation. For the preprocessing shown in \autoref{figure:dataset}, refer to \autoref{app:ecg_preprocessing}.

The Spiking Heidelberg dataset (\textbf{SHD}) is a benchmark audio-to-spike dataset specifically generated for SNNs \citep{cramer2020heidelberg}. It contains 10,420 recordings of spoken digits from 0 to 9 in English and German. The recordings were made with high fidelity and then 
converted into spike trains for 700 channels through a precise artificial inner ear model \citep{cramer2020heidelberg}. The spike trains were further processed with a discrete time scale of 4e-3\,s to obtain sequence length of 250 with zero-padding. This preprocessing was conducted to make the dataset comparable to \citet{yin2021accurate}. We used 7,341 sequences for training, 815 for validation, and 2,264 for inference.  

\section{Results}

\begin{table*}[t]
    \footnotesize
    \centering
    \caption{\label{tab:rsnn} Results between vanilla RF, BRF, BHRF, ALIF \citep{yin2021accurate}, and other \textit{SoTa} models: DCLS-Delays (DCLS-D) \cite{hammouamri2023learning} and RadLIF \cite{bittar2022surrogate} on various sequential classification tasks. Architecture represents the number of neurons in each layer. 
    The average accuracy over five runs shown for the RF, BRF and BHRF models, excluding PS-MNIST BHRF. SOPs: average spike operations. \textsuperscript{*}\textit{RSNN-SoTa} \textsuperscript{**}\textit{SNN-SoTa}}
    \begin{tabularx}{\textwidth}{C c l c l c c c}
    \toprule
    \textbf{Task} & \textbf{Model} & \textbf{Architecture} & \textbf{No. params. ($\downarrow$)}  & \textbf{Test Acc. ($\uparrow$)} & \textbf{SOPs ($\downarrow$)} & \textbf{SOPs/step ($\downarrow$)} & \textbf{SOP Ratio ($\downarrow$)}\\
    \midrule
     \multirow{4}{*}{\small S-MNIST} &\small ALIF\textsuperscript{*} & 4,64,$(256)^{2}$,10 & 156,126 & 98.7\,\% & 70,810.8 & 90.32 & 1.00$\times$\\
    & RF & 1,256,10 & \bf68,874 & 98.0$\pm 0.4 \, \%$&  29,034.8 & 37.03 & 0.41$\times$\\
    & \small BRF  & 1,256,10 & \bf68,874 & 99.0$\pm 0.1\,\%$& \bf15,462.6 & \bf19.72 & 0.22$\times$\\
    &\small BHRF & 1,256,10 & \bf68,874 & \bf{99.1$\pm 0.1\,\%$}& 21,565.7 & 25.51 & 0.30$\times$\\
    \midrule
    \multirow{4}{*}{\small PS-MNIST} &\small ALIF\textsuperscript{*}  & 4,64,$(256)^{2}$,10 & 156,126 &  94.3\,\%& 59,772.1 & 76.24 & 1.00$\times$\\
    & RF & 1,256,10 & \bf68,874 & 9.9$\pm 0.8\,\%$&  66,474.2 & 84.79 & 1.11$\times$\\
    & BRF  & 1,256,10 & \bf68,874 & 95.0$\pm 0.2\,\%$& 27,839.7 & 35.51 & 0.46$\times$\\
    & \small BHRF& 1,256,10 & \bf68,874 & \bf95.2\,\% & \bf24,564.2 & \bf33.33 & 0.41$\times$\\
    \midrule
    \multirow{3}{*}{\small ECG} & \small ALIF\textsuperscript{*} & 4,36,6 & 1,776 & 85.9\,\% & 35,011.2 & 26.93 & 1.00$\times$\\ 
    & RF & 4,36,6 & \bf1,734 & 85.5$\pm 0.7\,\%$& 11981.9 & 9.22 & 0.34$\times$\\
    & \small BRF & 4,36,6 & \bf1,734 & 85.8$\pm 0.7\,\%$& 6,307.7 & 4.85 & 0.18$\times$\\
    & \small BHRF & 4,36,6 & \bf1,734 & \bf87.0$\pm 0.4\,\%$& \bf6,233.8& \bf4.80 & 0.18$\times$\\
    \midrule
    \multirow{6}{*}{\small SHD} & DCLS-D\textsuperscript{**} & 700,$(256)^{2}$,20 & $\approx$200,000 & \bf95.1$\pm0.2\,\%$ & - & - & - \\
    & RadLIF\textsuperscript{*} & 700,$(1024)^{3}$,20 & 3,893,288 & 94.62$\%$ & - & - & - \\
    \cdashline{2-8}[2pt/1pt]
    &\small ALIF & 700,$(128)^{2}$,20 & 142,120 & 90.4\,\% & 24,690.0 & 98.76 & 1.00$\times$\\
    & RF  & 700,128,20 & \bf108,820 & 89.2$\pm 0.6\,\%$&  4750.2 & 19.00 & 0.19$\times$\\
    &\small BRF   & 700,128,20 & \bf108,820 &91.7$\pm 0.8\,\%$  & \bf3,502.6  & \bf14.01 & 0.14$\times$\\
    & \small BHRF  & 700,128,20 & \bf108,820 & 92.7$\pm 0.7\,\%$&  4,139.5& 16.56 & 0.17$\times$\\
    \bottomrule
    \end{tabularx}
    \label{tab:overview_acc}
\end{table*}

The results, the network architecture, the number of parameters, and the number of produced spikes between the best-performing BRF-, RF- and BHRF-RSNN and the baseline ALIF-RSNNs (see \autoref{alif}) or other state-of-the-art (SoTa) models, are compared for each classification task in \autoref{tab:overview_acc}. The BRF- and BHRF-RSNN hyperparameters for each dataset are shown in \autoref{app:lc_hyp} in \autoref{sec:hyperparams}. The BHRF and BRF models outperformed the RSNN SoTa for four and three tasks respectively. Complementing accuracy, we also find much enhanced sparseness in the BRF and BHRF networks while using substantially fewer parameters compared to the ALIF networks. Note that RF model shown in \autoref{tab:overview_acc} did not include a reset. Adding either a hard or soft reset to the vanilla RF neuron dramatically affects its task performance as shown in \autoref{app:reset_plot} in \autoref{sec:reset}.

We also compare the theoretical energy efficiency of the RF and ALIF models by computing their respective total spiking operations (SOPs) (see \autoref{eq:sop_equation}) and SOPs per sequence step. As both measures were considerably smaller for the BRF and BHRF models, this implies they require less computation to achieve better or comparable performance. This may be due to the resonating spiking behavior of the neuron, needing fewer spikes to represent its frequency, as well as the refractory period and smooth reset that hinders continuous spiking. A particularly large difference was seen in the SHD task, where the BRF model spiked only 14\% of the ALIF model. 

Compared to the standard RF model without reset, our balanced variants achieved significantly sparser activity while obtaining higher performance. Especially for the PS-MNIST, the standard RF model stayed at chance, while our balanced RF variations outperformed RSNN-SoTA. This is due to the damping factor and angular frequency combination leading to a highly diverging behavior. It especially affects the PS-MNIST dataset, as it requires a wider range of resonant frequencies to counter the increased randomness in the signal. For example, neurons with $\omega=70$ rad/s and $b_c=-1$ leads to higher magnitude towards the end of the sequence. Having many of such neurons causes the whole system to diverge and become unstable. On a different note, the performance of the BRF model depends on the initialization of the parameters and it generally under-performs when the angular frequencies are too far from the frequencies underlying the dataset.

Overall, we find that RSNNs comprised of either BRF and BHRF neurons exceeded performance for all classification tasks compared to the baseline ALIF-RSNN. Compared to standard RF neurons, the smooth reset, refractory period, and divergence boundary significantly improved the stability and efficiency of the RF parameters, possibly exploiting the model's resonant properties. We study this in detail below. 

\begin{table}[t]
\small
\caption{Quantitative convergence results. The average number of epochs after which $95\,\%$, $98\,\%$, and $100\,\%$ of the final test accuracy on the learning curve was reached, respectively.}
    \centering
\begin{tabularx}{\linewidth}{c C c c c}
\toprule
\textbf{Task} & \textbf{Model} & \textbf{$\mathbf{95\,\%}$ ($\downarrow$)} & \textbf{$\mathbf{98\,\%}$ ($\downarrow$)} & \textbf{$\mathbf{100\,\%}$ ($\downarrow$)} \\
\midrule
 \multirow{4}{*}{S-MNIST}
 &\small ALIF &  105 & 162 & 276\\
  &\small RF &  134 & 172 & 263\\
& \small BRF  &  \textbf{3} & 29 & 246\\
 & \small BHRF &  5 & \textbf{12} & \textbf{119}\\
\midrule
\multirow{3}{*}{PS-MNIST} & \small ALIF & 143 & 200 & 265\\
& \small BRF   & 10 & 39 & 282\\
 & \small BHRF & \textbf{6} & \textbf{19} & \textbf{123} \\
\midrule
\multirow{4}{*}{ECG} & \small ALIF & 51 & 157 & 282\\
 &\small RF &  8 & \textbf{29} & 112\\
& \small BRF  & \textbf{6} & 32 & \textbf{75}\\
& \small BHRF & 15 & 38 & 109\\
\midrule
\multirow{3}{*}{SHD}&\small ALIF & 14 & 15 & 16\\
 &\small RF &  \textbf{2} & 4 & \textbf{5}\\
&\small BRF    & \textbf{2} & \textbf{3} & 7\\
& \small BHRF  & 3 & 5 & 8\\
\bottomrule
\end{tabularx}
    \label{tab:final_percent}
\end{table}

\paragraph{Convergence.}

\begin{figure*}[t]
    \centering
    \includegraphics[width=\linewidth]{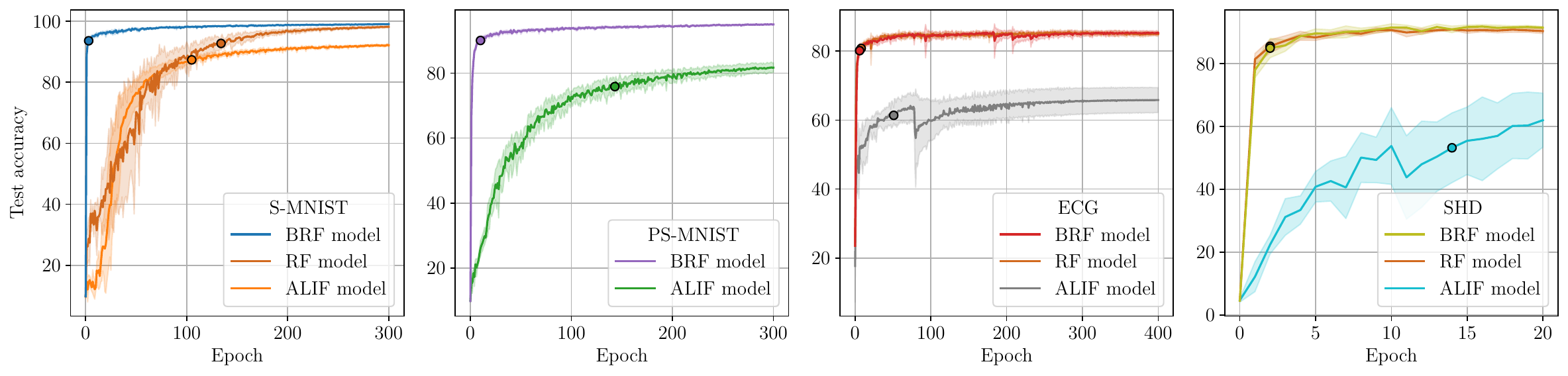}
    \includegraphics[width=\linewidth]{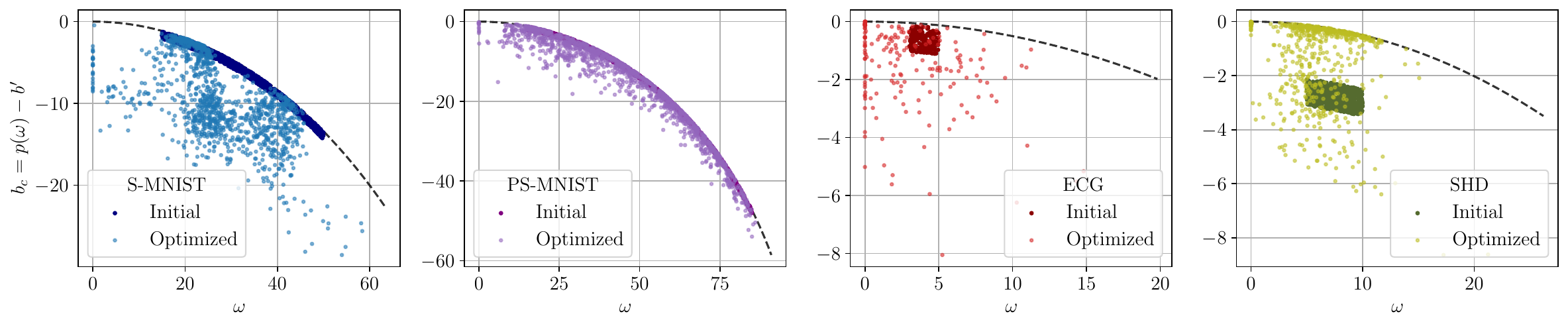}
    \caption{Top row: S-MNIST, PS-MNIST, ECG, and SHD learning curve between BRF, RF and ALIF model. Each curve averaged per epoch (solid line) with standard deviation (shaded area) over five runs. The dot on the accuracy curves depict the point at which 95\,\% of the final accuracy was reached. Bottom row: S-MNIST, PS-MNIST, ECG, and SHD initial and optimized BRF parameter combinations: angular frequency $\omega$ and b-offset $b'$ for all runs. Dashed line is the divergence boundary. RF models are simulated without reset. For convergence with traditional reset see \autoref{sec:reset}.}
    \label{figure:smnist_lc}
\end{figure*}

The RF, BRF, and ALIF model learning curves are presented in \autoref{figure:smnist_lc}. We reduced the ALIF model to the same size and number of trainable parameters as the BRF model. For the S-MNIST and PS-MNIST tasks, the ALIF networks only learn effectively when the truncated BPTT (TBPTT) with truncation step 50 is applied. Nonetheless, the fully back-propagated BRF networks converge significantly faster than the TBPTT ALIF networks, also evident from the quantitative convergence results (\autoref{tab:final_percent}).

A similar pattern of fast and stable convergence is observed for the ECG dataset and SHD. The convergence between RF and BRF for SHD and ECG are similar because the vanilla RF neurons resonate with lower frequencies, affected less by diverging behavior. Note that the RF model shown in \autoref{figure:smnist_lc} did not include a reset. Adding either soft or hard reset leads to slower and more unstable learning, as shown in \autoref{app:reset_plot} in \autoref{sec:reset}. 
\autoref{figure:smnist_lc} highlights the variation in training of RF networks depending on the dataset, which is alleviated by the BRF network.

While numerically large, the learning rates of the (B)RF-RSNNs led to stable and fast convergence for most datasets. This is in contrast to standard SNNs where large learning rates lead to poor performance.  
Here, the double-Gaussian surrogate gradient function \citep{yin2021accurate}  we used may have contributed by ensuring gradient flow even for neurons far from firing. 
Combined with the high learning rate, this may have enabled non-spiking neurons to effectively shift their angular frequencies.

The ECG task was learned most effectively for the BRF and BHRF networks with a small batch size and high learning rate. A smaller batch size can better capture detailed variations within the dataset, notably the timing of the input spikes, which may have been important for per-step wave classification.

Ongoing investigations \cite{higuchi2024convergence} suggest that the shape of the error landscape may be a major contributor to the fast and stable convergence of the BRF model (see \autoref{figure:error-landscape} in \autoref{sec:error}).

\paragraph{Parameter analysis. }
We also compared the trainable BRF parameters $\omega$ and $b'$ between the initial and trained networks to obtain an intuition of the functioning of the neuron, as shown in \autoref{figure:smnist_lc}. 
We see that optimization substantially shifted the parameters, demonstrating that the gradients in the B(H)RF networks were effectively propagated. 

For the S-MNIST task, plotting $\omega$-$b_c$ (\autoref{figure:smnist_lc}) shows clusters around $\omega$ values of 18 to 28 $rad/s$ 
and around $\omega$ of 40 $rad/s$ with a wide range of $b_c$. The dampening factor close to zero indicates near-sustained oscillation behavior, preserving resonance amplitude for an extended duration. 
The clustered variety in $b_c$ values suggests that short- and long-term memory are essential for effective learning of the S-MNIST task in B(H)RF networks. Additionally, 
the distribution of $\omega$ for S-MNIST task showed a bimodal distribution with the highest peak at 23.3 to 25.3 $rad/s$ and the second peak at 40.8 to 42.8 $rad/s$. 

The MNIST digits all contain some form of diagonal line with a width of 3 to 6 pixels, most prominently observed in digit 1. When converted into a sequence row-by-row, the lines are distributed among the sequence with a periodic signal of about 25 pixels. Considering the pixels as discrete time steps of $\delta$ = 0.01, the theoretical period T frequently observed in the signal is T = 0.25 $s$. Thus, the most frequent theoretical angular frequency $\omega'$ underlying the signal is around
$\omega' = 2 \pi / 0.25\,s = 25.13\,rad/s$
which is close to the most frequently learned angular frequencies of the BRF neuron. 
This simple calculation thus suggests that the BRF neurons were able to learn meaningful frequencies that underlie the S-MNIST dataset. For the PS-MNIST, we see more spread of $\omega$ and $b_c$ parameters, as the specific characteristic frequency structure from S-MNIST is essentially obscured by the permutation.

Inspecting spike train ECG samples (\autoref{figure:dataset}), we observe distinguishable periodic spikes with 300 time step intervals, corresponding to a theoretical angular frequency of $\omega' = 2.09$ $rad/s$, and indeed a cluster at about 2 $rad/s$ is present in the $\omega$-$b_c$ plot. 
The theoretical and optimized angular frequencies align, demonstrating learning of the dynamical signal by the network.

\begin{figure}[t]
    \centering
    \includegraphics[width=0.9\linewidth]{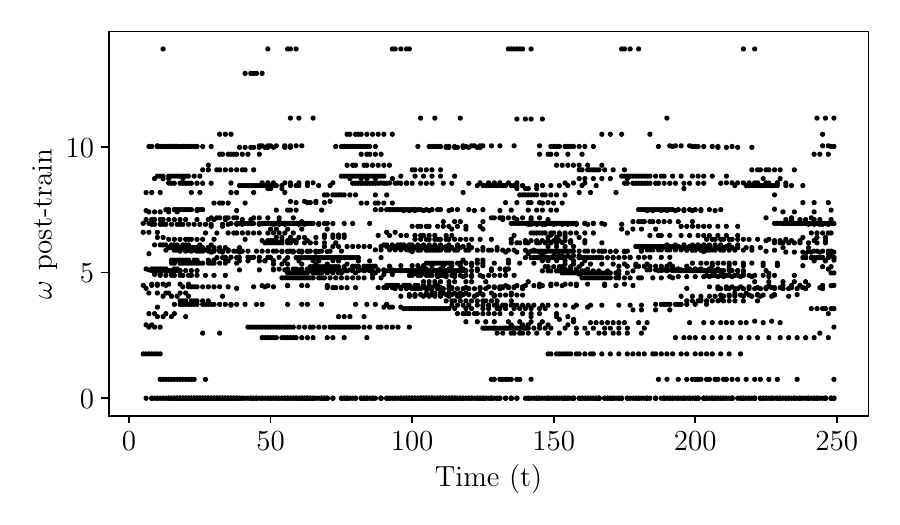}
    \caption{Raster plot of network activity for an SHD sample with label 10  after training. The BRF neurons are sorted by their angular frequency. Black dots denote output spikes.}
    \label{figure:shd_raster}
\end{figure}

\paragraph{Weight Sparsity. }
\begin{figure}[t]
    \centering
    \includegraphics[width=\linewidth]{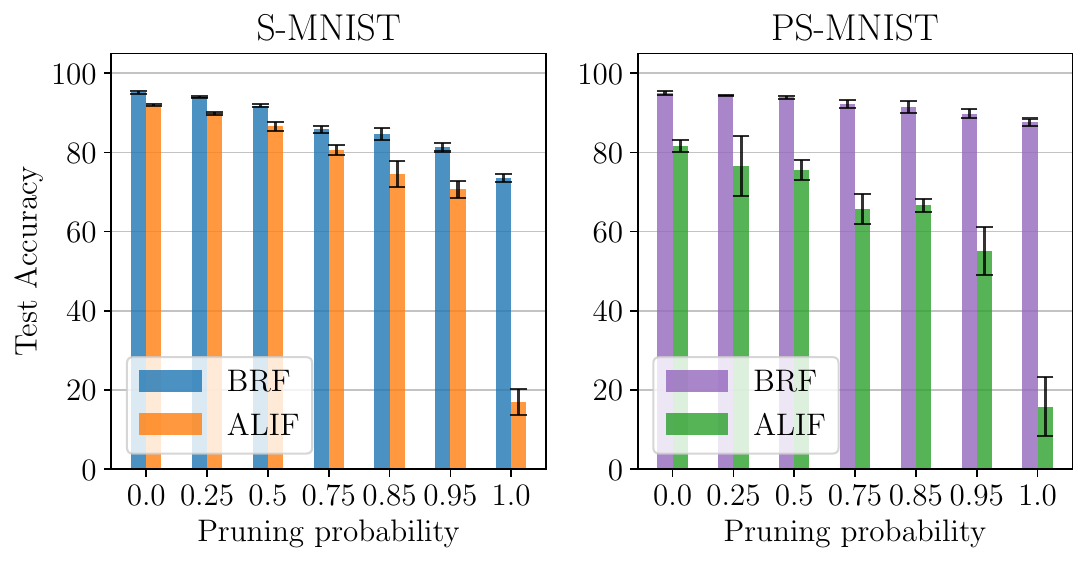}
    \caption{\small Test accuracy over pruning probability for (a) S-MNIST and (b) PS-MNIST. 1.0 pruning indicate no recurrent weights.}
    \label{figure:weight_sparsity}
\end{figure}

We explored the effect of pruning recurrent weights in an ablation study for the S-MNIST and PS-MNIST tasks between comparable BRF and ALIF networks, as detailed in \autoref{app:smnist_lc_hyp} and results are shown in  \autoref{figure:weight_sparsity}. While the BRF neuron provides consistent performance throughout pruning, the ALIF model performs poorly when lacking recurrent weights. This demonstrates that explicit recurrencies are not necessary for the BRF networks with resonating properties to learn reasonably  well, while these connections are crucial for the ALIF networks. Moreover, the BRF networks' accuracy has a smaller standard deviation throughout compared to the ALIF networks, also showing that the BRF network performance is not affected by the specific pruned weights as much as the ALIF networks.

\section{Discussion}

The results from our BRF-RSNN implementation lead to considerable insight into the workings of the RF neurons. Importantly, the learning curve and parameter analyses of the datasets show that the BRF networks learn meaningful parameters by favorably tuning the resonating behavior, modeling the complex dynamics not yet observed in large-scale simulations. We observe favorable oscillatory behavior of the membrane potential and increasing frequency of periodic spikes with higher angular frequency in the raster plot for all datasets. For SHD, output spikes are present towards the end of the sequence, even with the zero padding of the input signal. As most signals are still periodic, it suggests controlled oscillation and spiking that maintains relevant frequencies until the end of the sequence. 

Note that we conducted the experiments without biases for all RF models to investigate how many spikes are needed to keep the network dynamics alive and to solve the tasks. As a result, a few of the BRF neurons in the network learned to fire consistently throughout the sequence as seen in \autoref{figure:shd_raster} and thus effectively ``imitated'' a bias injecting constant current into the network. In contrast, ALIF networks need explicit biases to perform well. The BRF networks are thus more flexible in modeling behavior that cannot be learned by the integrator neurons alone.

The neurons learn their own favored frequencies and, consequently, focus on individual time scales. This can be loosely seen as a homeostatic regulatory process maximizing local information gain in the realms of distributed, collective task solving within the recurrent networks. 

Further exploration of the post-trained models against various noise types and numerical quantization \citep{stromatias2015robustness, park2021noise}, shows consistently high robustness of the BRF networks compared to the ALIF networks, detailed in \autoref{sec:noise-robustness}, \autoref{app:noise_robust}. The BRF networks are slightly more robust than the RF networks for some noise variants, despite the constant low SOP of the BRF networks. Hardware implementations are prone to such noises \citep{stromatias2015robustness}, which suggests the BRF neurons to be suitable for hardware applications. We further aim to realize inference with these neurons on Loihi 2, which provides support for RF neuron variations \citep{davies2018loihi, shrestha2023efficient}.

Concerning the network dimension, we found that adding additional resonator neurons into the single hidden layer beyond a certain capacity did not significantly improve the performance further.

One drawback of our BRF- and BHRF-RSNNs is the difficulty in finding an optimal initial parameterization of the angular frequencies and the dampening factor offset, since both determine the models' performance. The analyses of the parameters suggest the possibility of applying the Fourier transform on the datasets a-priori to approximate the range of angular frequencies the BRF neurons require to optimally learn the data.

It should be noted that the computational complexity of BRF-RSNNs is similar to single-layered ALIF models. However, due to the faster convergence the training can be stopped much earlier, which effectively results in a drastically reduced training time. Implementation-wise, another speed-up can be achieved by using forward gradient injection \citep{otte2024fgi} for modeling the surrogate gradient functions in combination with automatic model optimization routines such as TorchScript in PyTorch (some results are presented in \autoref{sec:fgi}).

\section{Conclusion}

We introduced BRF- and BHRF-RSNNs as a principled resonating spiking neuron model that solves the training difficulties encountered for networks comprised of standard RF neurons while demonstrating the effectiveness of internal resonating state as a form of long-term memory. 
Our BRF- and BHRF-RSNNs with smaller network architectures and sparser spiking outperformed deep baseline models. The learning curves showed that the BRF networks converged much faster than the ALIF networks. Furthermore, the stable convergence hints at good reproducibility of the BRF model's performance. The BRF parameter analyses indicate the network to learn frequencies underlying the input signals and showed resilience against sparser recurrent connectivity compared to the ALIF network.

Possible future research may be to scale up the simulations to much larger and more complex problems, such as the CIFAR \cite{krizhevsky2009learning} or the Google Speech Command dataset \cite{warden2018speech}. It is of particular interest to apply the BRF model to raw audio processing tasks, because of its ability to extract periodic pattern within the time domain. 

Another line of research could be to combine BRF neurons with online learning approaches, such as e-prop \citep{bellec2020solution}, which would broaden its range of applications due to memory efficiency and faster processing of longer sequences. 
Additionally, trainable refractory period decays for flexible adaptive thresholding could be introduced, as well as trainable simulated variable time constants to foster multi-temporal resolution processing.

This study marks an initial exploration to investigate the working of stable resonating neurons in the context of recurrent spiking neural network. Our models show comparable results and/or exceeds the state-of-the-art for RSNNs. We see many possible approaches to extending our work, as we kept the implementation to its simplest form. The presented work thus provides a foundation for further research of possible BRF-RSNN variants.  

\section*{Acknowledgements}
A part of the experiments were conducted using the ML Cloud infrastructure of the Machine Learning Cluster of Excellence (EXC number 2064/1 – Project number 390727645) at the University of Tübingen.

Sebastian Otte was supported by a Feodor Lynen fellowship of the Alexander von Humboldt Foundation.

\section*{Impact Statement}
The aim of this work is to advance the field of machine learning, with a specific focus on spiking neural networks (SNNs) to augment the effectiveness and energy efficiency of AI systems, making them more sustainable. It is worth noting that, like most developments in machine learning research, our work may have various societal consequences, but none of which we think needs special consideration here.

\bibliography{main}
\bibliographystyle{icml2024}

\newpage
\appendix
\onecolumn

\section{Appendix}

\subsection{Traditional soft and hard reset}\label{app:reset_mech}
The traditional soft and hard reset mechanisms were formulated respectively: 
\begin{equation}\label{eq:soft-hard} 
    u(t) =  u'(t) - (1 + i) \, z(t) \, \vartheta
\end{equation} 
\begin{equation}
    u(t) =  [1 - (1 + i) \, z(t)] \, u'(t)
\end{equation} 
for which $u'(t)$ represents the membrane potential before reset.

\subsection{Baseline ALIF neuron} \label{alif}
The formulation of the baseline ALIF neuron \cite{yin2021accurate} is as follows with $\vartheta = 0.01 $ and $\beta = 1.8$:
\begin{equation}
\vartheta^t = \vartheta + \beta \, a(t)
\end{equation}
\begin{equation}
a(t) = \rho \, a(t-\delta) + (1-\rho) \, z(t-\delta)
\end{equation}
\begin{equation}
u'(t) = \alpha \, u(t-\delta) + (1 - \alpha)  \, I(t)
\end{equation}
\begin{equation}
z(t) = \Theta \, (u'(t) - \vartheta^t)
\end{equation}
\begin{equation}
u(t) = u'(t) - z(t) \, \vartheta^t
\end{equation}
where $\rho = e^{-\frac{\delta}{\tau_{a}}}\in (0,1)$ is the adaptive threshold decay constant, $\tau_{a}$ the time constant and $\alpha = e^{-\frac{\delta}{\tau_{m}}}\in (0,1)$ the membrane potential decay constant. $a(t)$ is the accumulative activity of the spiking behavior of the neuron. The membrane potential $u(t)$ is soft reset with the adaptive threshold $\vartheta^t$ when the neuron fires ($z(t)=1$).

\subsection{Explicit form of the Izhikevich RF neuron}\label{app:explicit_subthreshold}
We can reform the membrane equation to $u(t) = (1 + \delta \,(b + i \, \omega)) \, u(t-\delta) + \delta \, I(t)$.
Consider discrete time step $\delta$ with t = 0, $\delta$, $2\delta$, $3\delta, \,, T\delta$. Assume a spike injected only at $t=\delta$ with $I(\delta) = 1$ and initial membrane potential $u(0) = 0$, then: 
\begin{align}
    u(\delta) &= \delta\\
    u(2\delta) &= \delta \, (1  + \delta\, (b + i \, \omega)) \\
    u(3\delta) &= (1 + \delta \,(b + i \, \omega))(\delta \, (1  + \delta\, (b + i \, \omega)))\\
    &= \delta \, (1  + \delta\, (b + i \, \omega))^2 \\
    u(4\delta) &= (1 + \delta \,(b + i \, \omega)) (\delta \, (1  + \delta\, (b + i \, \omega))^2)\\
    &= \delta \, (1  + \delta\, (b + i \, \omega))^3\\
    \cdots \\
    u(t) &= \delta \, (1  + \delta\, (b + i \, \omega))^{\frac{t}{\delta}-1}
\end{align}

\subsection{Frequency response plot generation}\label{app:freq_method}
The subthreshold responses were explored with randomly chosen angular frequency $\omega \in [0,100)$ and $b' \in (0,10)$ for the RF neurons. Spiking input signals with frequencies relative to the angular frequencies of $\{0.1, 0.2 \cdots, 100\}$ were input to the neurons over 20 seconds with a discrete-time scale of $\delta = 0.001$. The positive spikes were input in-phase of the period corresponding to the angular frequency: 
\begin{equation}
T = \frac{2\pi}{\text{angular frequency}}
\end{equation}
The mean absolute magnitude of the membrane potential over the whole sequence was calculated and plotted to get the response of the neuron per tested frequency signal. Similarly conducted for the HRF neurons.

\begin{figure}[h]
    \centering
    \includegraphics[width=0.58\linewidth]{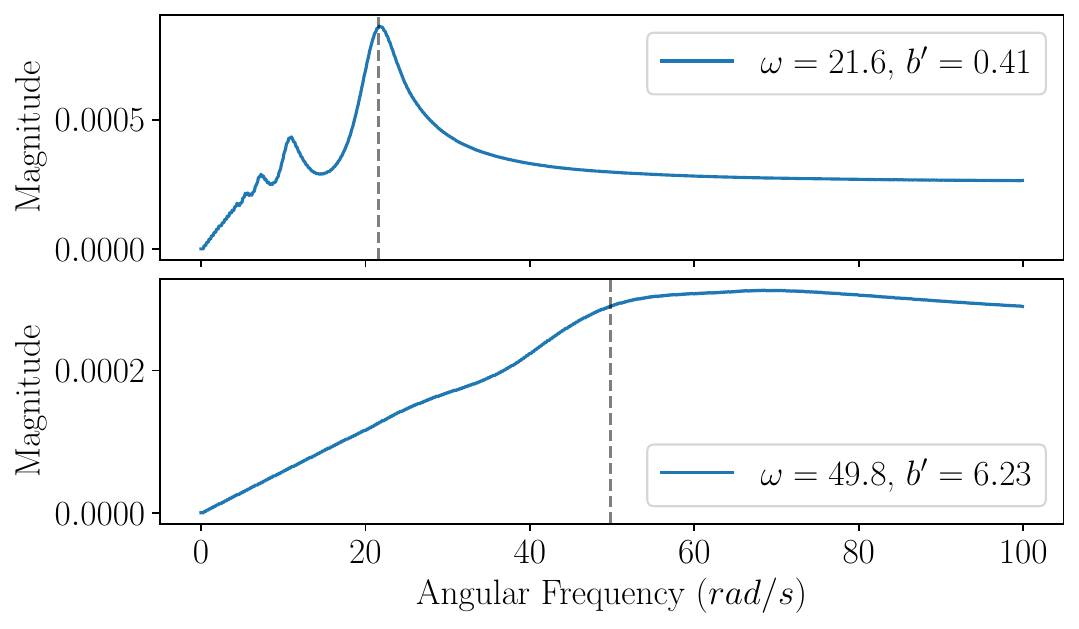}
    \caption{HRF neuron frequency response plots for exemplary omega $\omega$ and b-offset $b'$ combinations with $\delta = 0.001$.}
    \label{app:freq_plot_bhrf}
\end{figure}

\subsection{Divergence Boundary}\label{app:detla_div_bd}
The discrete-time scale $\delta$ has a large impact on the divergence behavior. Figure below shows the curve of divergence boundary for time scales above and below the default $\delta =0.01$. The membrane potential converges if the combination of $\omega$ and $b_c$ is below the line. When considering the $\omega$ range of 0 to 100 $rad/s$, the slope of the sustained oscillation curve becomes flatter with smaller $\delta$. The more precise the oscillation is modeled, the less likely the system will diverge. Indeed, the discrete approximation of the continuous differential equation becomes more precise by computing smaller time scales.

\begin{figure}[h]
    \centering
    \includegraphics[width=0.6\linewidth]{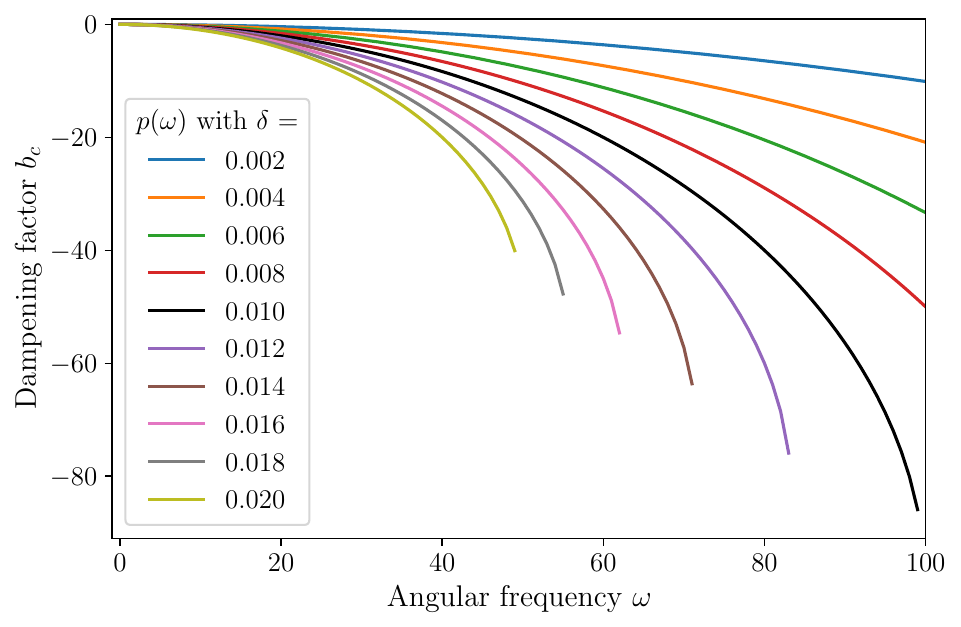}
    \caption{Divergence boundary depending on $\delta \in \{0.002,\cdots, 0.02\}$}
    \label{fig:delta_div_bd}
\end{figure}

\subsection{Experimental Setup} \label{app:network}

The fully recurrent RF-RSNN with $m$ input, $h$ hidden BRF or BHRF, and $C$ leaky integrator (LI) output neurons was computed using PyTorch \citep{paszke2017automaticpytorch}. The backpropagation through time (BPTT) algorithm with minibatch size $B$ was used with the Adam \citep{kingma2014adam}, RAdam \cite{liu2019radam}, or the RMSprop \cite{hinton2012rmsprop} depending on the network and task. The input from the current sequence step $\mathbf{x}^t \in \mathbb{M}_{B \times m}$ and the output spiking of the recurrent hidden neurons from the previous time step $\mathbf{z}^{t-1} \in \mathbb{M}_{B \times h}$ were combined by implementing a fully connected linear layer. This combined signal $\mathbf{x}^t \in \mathbb{M}^{B \times h}$ corresponded to the familiar injected current $I(t)$ but was written in matrix form to represent all hidden neurons in the minibatch. Although not explicit in the Algorithms, $\mathbf{w_{in, rec}} \in \mathbb{M}_{h \times (m+h)}$ and $\mathbf{w_{out}} \in \mathbb{M}_{C \times h}$ represent the strength of the connections between the input-to-hidden, hidden-to-hidden, and hidden-to-output neurons, which were optimized. No biases were trained or used in the network. The B(H)RF neurons with $\boldsymbol{\omega}$ and $\mathbf{b'} \in \mathbb{R}^h$ were updated over the time sequence of $B$ data samples with $\mathbf{u}^t, (\mathbf{v}^t), \mathbf{b}^t, \mathbf{q}^t, \boldsymbol{\vartheta}^t \in \mathbb{M}_{B \times h}$. The LI membrane potential time decay constant $\boldsymbol{\tau_{m,out}}\in \mathbb{R}_{> 0}^C $ was also learned. The negative log-likelihood (NLL) or the cross-entropy (CE) loss was implemented as our criterion depending on the simulations. 

For the ALIF implementation, the adaptive threshold time decay constant $\boldsymbol{\tau_a} \in \mathbb{R}_{> 0}^h$ is additionally learned with the membrane potential time decay constant $\boldsymbol{\tau_m} \in \mathbb{R}_{> 0}^h$.

In the case of average sequence loss with NLL, the logarithmic softmax function was computed for all output neurons per time step $\mathbf{\hat{y}}^t = log(\text{softmax}(\mathbf{u}^t_{out}))$, which was further propagated into the loss function: $\mathcal{L}^t =\text{criterion}(\mathbf{\hat{y}}^t,\mathbf{y}^t) = \frac{1}{B}\sum_{c=1}^{C} - y^t_c \, \hat{y}^t_c$ where $\mathbf{y}^t \in \mathbb{M}_{B \times C}$ is the target label at time step $t$ represented as one-hot vectors. The mean loss was computed as  $\mathcal{L} = \frac{1}{T} \sum^T_{t=1} \mathcal{L}^t$ with sequence length T. In the case of label-last loss, only the loss from the last time step was propagated backwards. 

The backward pass of the BPTT algorithm was internally computed with the automatic differentiation engine in PyTorch \citep{paszke2017automaticpytorch}. For the non-differentiable Heaviside function in the B(H)RF and ALIF neurons, the multi-Gaussian function \cite{yin2021accurate} was manually implemented as the surrogate gradient \cite{neftci2019surrogate}:
\begin{equation}\label{eq:multi-gaussian}
\frac{\partial \Theta}{\partial u}
=_{def} (1 + h) g(u, \vartheta, \sigma) - 2h
g(u, \vartheta, s\sigma)
\end{equation}
with
\begin{equation}
    g(x, \mu, \sigma) = e^{-\frac{1}{2}\left(\frac{x - \mu}{\sigma}\right)^{2}}
\end{equation}
with $h{=}0.15$, $s{=}6$, $\sigma{=}0.5$ similar to \citeauthor{yin2021accurate} \yrcite{yin2021accurate}. The negative gradient values were motivated by the leaky rectified and exponential linear unit that prevented the ``dying-ReLU" problem \citep{lu2019dying}. With these values, the multi-Gaussian surrogate gradient function promoted the neurons to spike despite an initial low membrane potential, which contributed to effective learning. 

The best model was the model saved with the highest average test loss over the first five runs by early stopping on the validation loss. The loss and accuracy of the train, validation, and test set were logged into Tensorboard \citep{abadi2016tensorflow} for analysis of the learning pattern. After obtaining the best model, the spiking operations (SOPs) were computed based on the five saved models by taking the total number of output spikes over the whole dataset $z_{sum}$ with respect to the number of data samples $N$: 
\begin{equation} \label{eq:sop_equation}
\operatorname{SOPs} = \frac{z_{sum}}{N}
\end{equation}
The results were compared with the baseline ALIF-RSNN.

For simulating the models and performing the experiments, we used multiple systems with different deep learning accelerators including NVIDIA GeForce RTX 2060, NVIDIA GeForce RTX 2080 Ti, NVIDIA GeForce RTX 3090, and NVIDIA A100, with PyTorch 2.0.1 on Python 3.10.4 and CUDA 11.7.

\subsection{ECG-QT database preprocessing} \label{app:ecg_preprocessing}
The preprocessed QT data were extracted from \citeauthor{yin2021accurate} \yrcite{yin2021accurate} to gain comparable results. The original sequences were segmented into smaller intervals, each containing 1300 ms of the recordings. Then, the two ECG signals were normalized and encoded into two separate spike trains via level-cross encoding. The positive gradients above and negative gradients below the threshold led to spikes with the threshold $L =  0.3$:

\begin{align}
s_+ &=
\begin{cases}
    1 & \text{if } x_t - x_{t-1} \geq L \\
    0 & \text{otherwise}
\end{cases} \\
 s_- &=
\begin{cases}
    1 & \text{if } x_t - x_{t-1} \leq -L \\
    0 & \text{otherwise}
\end{cases} 
\end{align}

The target labels were segmented correspondingly and predicted for each time step. \autoref{figure:dataset} color: blue, red, green, cyan, olive and purple corresponds to the label: P, PQ, QR, RS, ST. Five hundred fifty-seven segments were used for training, 61 for validation, and 141 for testing.

\subsection{Effect of refractory period and smooth reset} \label{sec:sop_rf_smr}
Refractory period (RP) and smooth reset (SmR) both result in reduced spiking operations, with their effects varying depending on the dataset. The SOP of the BRF neuron consistently yields the lowest SOP. It suggests the flexibility of the BRF neuron to learn various types of data efficiently, compared to the RF neuron. ECG optimized with RP spikes less than with SmR, whereas SHD spikes more with SmR; however even more efficient in both cases is the BRF with the combination of RP and SmR. PS-MNIST diverges with: no reset, RP, SmR, and combination of RP and SmR. Thus, it shows significantly high SOP for these variants. It also highlights the stability brought out by the divergence boundary (DB). S-MNIST trained with RP results in unstable performance and SOP.

\begin{figure}[h]
    \centering
    \includegraphics[width=\linewidth]{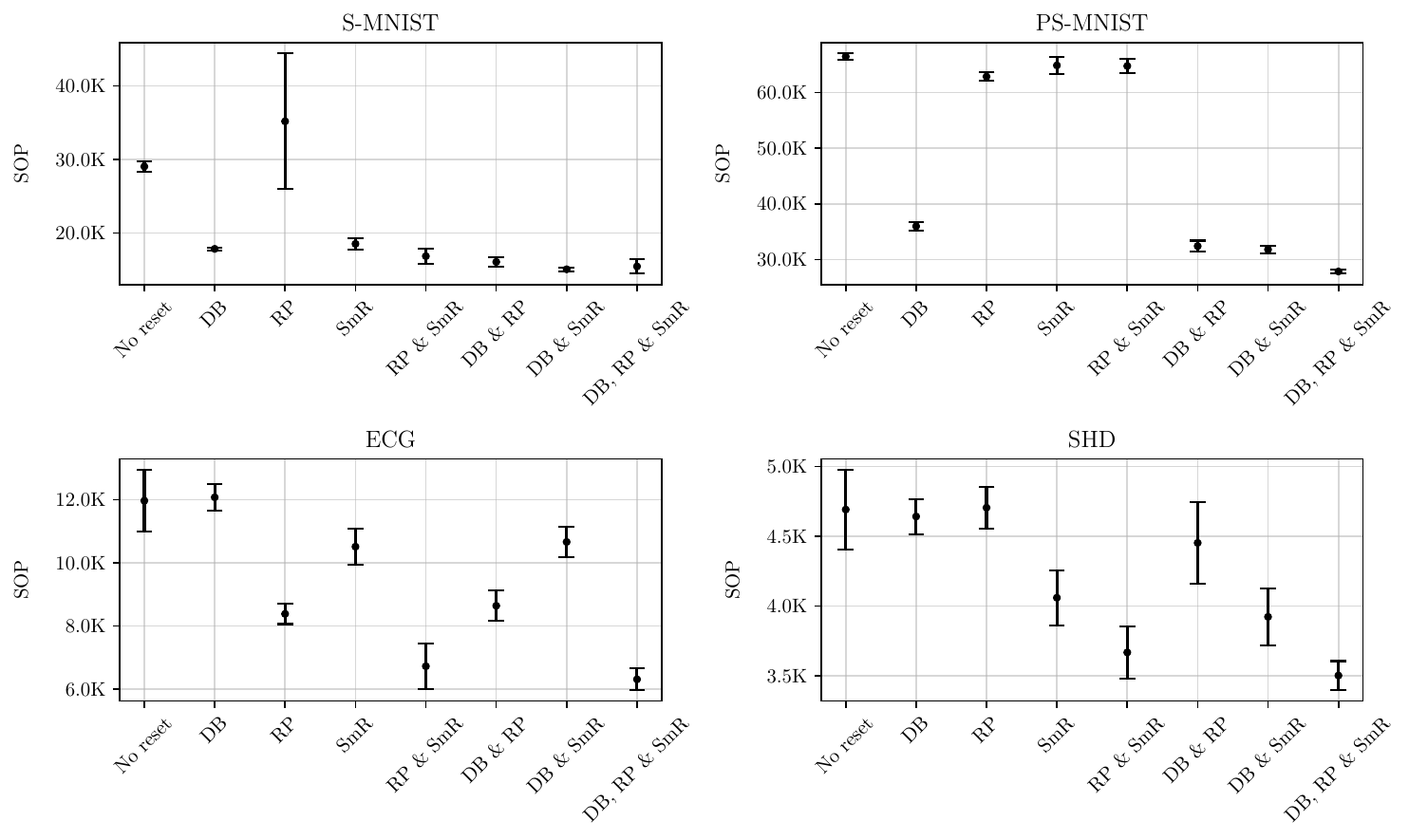}
    \caption{SOP result in RF networks with divergence boundary (DB), refractory period (RP), and smooth reset (SmR). Mean SOP over 5 runs plotted with standard deviation. DB, RP \& SmR denote the complete BRF neuron.}
    \label{app:rf_sp_sop_plot}
\end{figure}

\subsection{Reset variation and performance}\label{sec:reset}
\begin{figure}[h]
    \centering
    \includegraphics[width=0.85\linewidth]{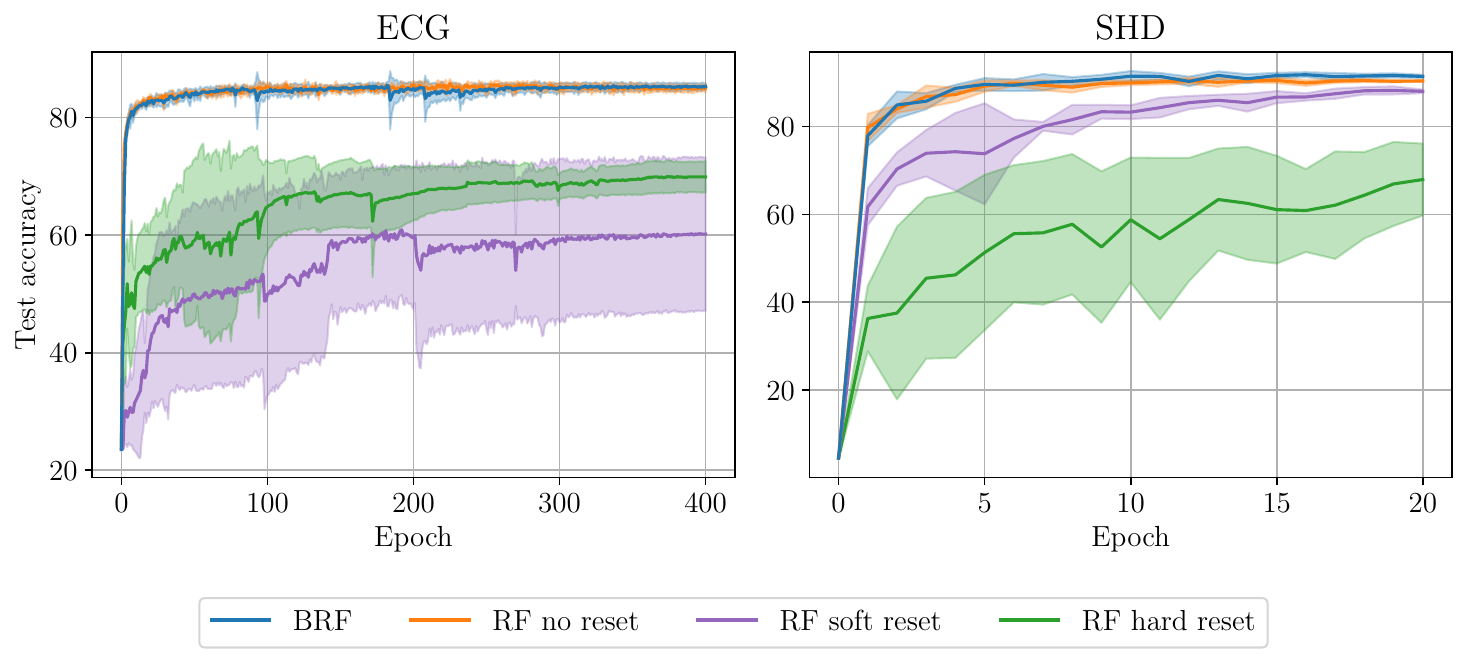}
    \caption{Convergence comparison of ECG and SHD over RF reset mechanisms. }
    \label{app:reset_plot}
\end{figure}

To explore the impact of the standard reset mechanisms (mentioned in \autoref{app:reset_mech}) on the resonator network, RF models with smooth, soft, and hard reset are optimized for ECG and SHD, as shown in \autoref{app:reset_plot}. Note that the traditional reset mechanisms applied to S-MNIST and PS-MNIST failed to converge. The figures show performance drop and slower convergence with the soft and hard reset, presumably due to the altered phase of the oscillation and the difficulty in continuous resonant spiking, as spikes may occur with twice the period of the original signal. Further facilitated by the result without reset, in which the convergences are on-par with the BRF neurons where the phase itself is preserved. As mentioned in the main text, the system did not diverge for the RF without reset for ECG and SHD in particular, due to the small angular frequencies leading to smaller error accumulation over time.


\subsection{Noise robustness} \label{sec:noise-robustness}
\begin{figure}[t!]
    \centering
    \includegraphics[width=0.95\linewidth]{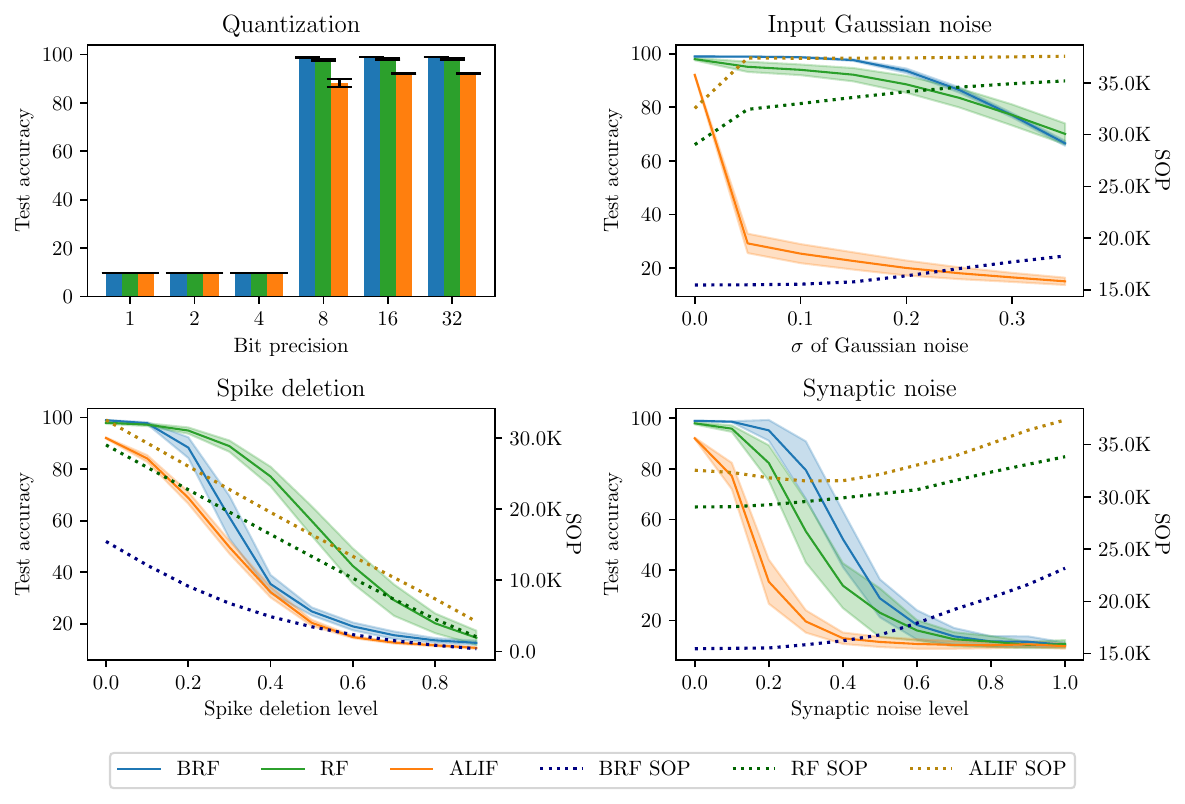}
    \caption{S-MNIST noise robustness studies conducted on saved best performing models. Solid line denote performance and dotted line the spiking operations (SOP) for each network.}
    \label{app:noise_robust}
\end{figure}

Noise robustness of post-trained models are further investigated with the methods applied by \citet{stromatias2015robustness} and \citet{park2021noise}. Hardware implementations of SNNs pose limitations to the performance due to restricted precision or noise in the signal. Here, four different type of restrictions and noise are explored.
Quantization reduces the bit precision representation (\autoref{app:noise_robust} top left). Noise in the input signal from the dataset are simulated with Gaussian noise with increasing standard deviation (\autoref{app:noise_robust} top right). For spike deletion \cite{park2021noise} level, a percentage of the spikes are removed from the network at each sequence step (\autoref{app:noise_robust} bottom left). 
Synaptic noise added Gaussian noise to the weights.

Overall the vanilla RF network without reset and the BRF network performed better when faced with various noise than the ALIF network. Especially for the noise introduced in the input, the RF and BRF network maintained their performance, while the ALIF network failed already with a small standard deviation in noise. For the BRF network, this is achieved despite the high spiking sparsity of the BRF network compared to the RF or ALIF network. The RF network is more robust against spike deletion than the BRF or ALIF network. This may be an effect of redundant spiking for the RF networks, which highlights a trade-off of spike deletion robustness and high spiking sparsity.

\subsection{Error landscape} \label{sec:error}
BRF-RSNN landscape has smooth and convex-like structure, whereas landscape of RF- and especially, ALIF-RSNN, has a rough surface with a narrow valley. Such smooth landscape represent high generalization and straightforward optimization, accounting for the fast convergence. 

We further found that the spectral radius of the membrane state transition mapping computed over one discrete time step results in unity or below by implementing the divergence boundary, thus effectively stabilizing the gradient and preserving its magnitude \citep{higuchi2024convergence}. This indicates that the divergence boundary contributes to fast and smooth learning.

\begin{figure}[h!]
    \centering
    \includegraphics[width=\linewidth]{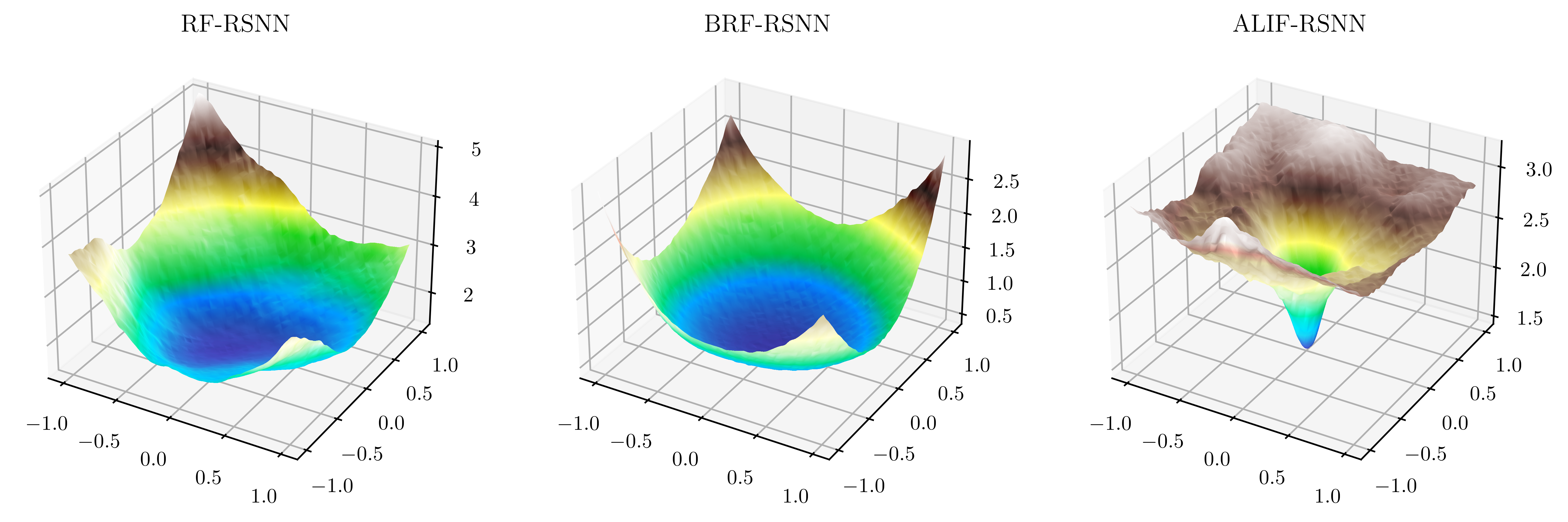}\\
    \includegraphics[width=\linewidth]{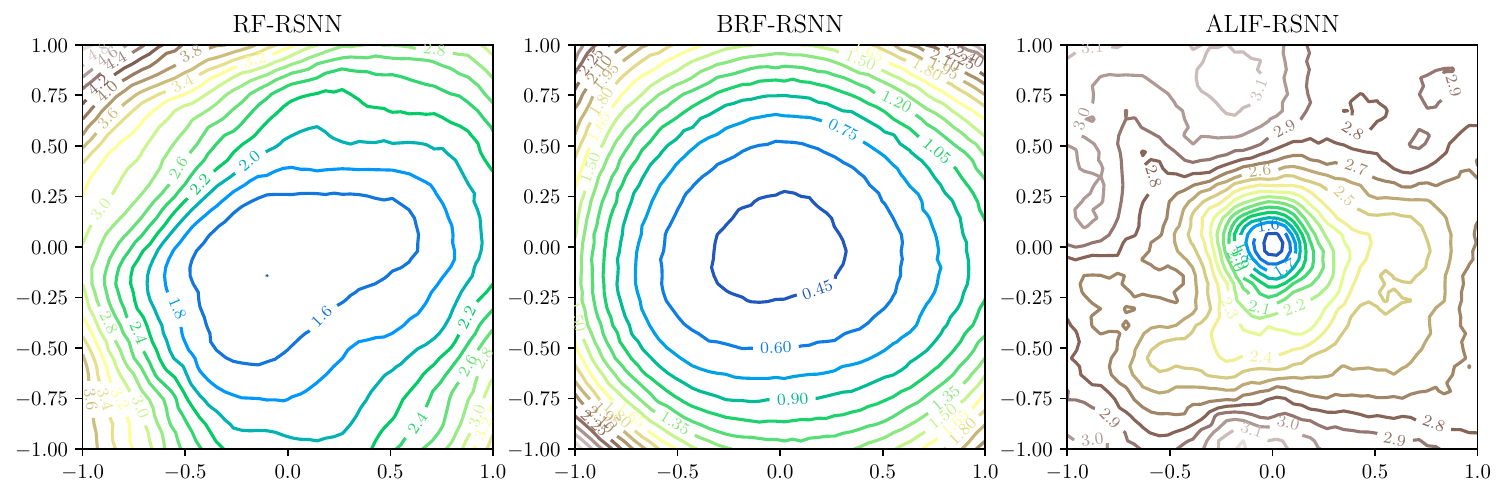}
    \caption{Error landscape plots for RF, BRF, and ALIF network on the S-MNIST dataset adopted from \citet{higuchi2024convergence}. Top: error surface plots. x and y axis correspond to $\alpha$ and $\beta$, i.e. parameter deviations, and the z axis to $f(\alpha, \beta)$, i.e. error \citep{li2018visualizing}. Bottom: error contour plots. Note that the value range and corresponding coloring differ between the diagrams to enhance visualization. }
    \label{figure:error-landscape}
\end{figure}

\pagebreak

\subsection{Hyperparameters}\label{sec:hyperparams}

\begin{table}[h!]
\small
\caption{Hyperparameters applied for the best-performing BRF, HBRF, and ALIF model with label-last loss (S-MNIST) and average sequence loss (PS-MNIST, ECG, SHD) and truncation step of 50 for TBPTT (in pruning). Same hyperparameters applied for BRF- and RF-BPTT models, where $b=b'$ for RF-RSNN. 
}
\label{app:lc_hyp} 
\centering
\begin{tabularx}{\linewidth}{lCCcCC}
\toprule
\multirow{2}{*}{S-MNIST} & \multicolumn{2}{c}{BRF-RSNN} & BHRF-RSNN& \multicolumn{2}{c}{ALIF-RSNN}  \\
& BPTT & TBPTT & BPTT& BPTT & TBPTT \\
\midrule
Network & \multicolumn{2}{c}{1 + 256 (fully recurrent) + 10} & 1 + 256 (fully recurrent) + 10& \multicolumn{2}{c}{1 + 256 (fully recurrent) + 10} \\
Learning rate (Lr) & 0.1 & 0.006 & 0.1& \multicolumn{2}{c}{0.001} \\
Loss function & \multicolumn{2}{c}{NLL}  & CE& \multicolumn{2}{c}{NLL}\\
Minibatch size & \multicolumn{2}{c}{256} & 256& \multicolumn{2}{c}{256} \\
Lr scheduling &  \multicolumn{2}{c}{LinearLR}& LinearLR & \multicolumn{2}{c}{LinearLR} \\
Optimizer & \multicolumn{2}{c}{Adam}& RAdam & \multicolumn{2}{c}{Adam} \\
Epochs & \multicolumn{2}{c}{300} & 400&  \multicolumn{2}{c}{300}  \\
\begin{tabular}[c]{@{}l@{}}Parameter\\ ~~initialization\end{tabular} & \multicolumn{2}{c}{\begin{tabular}[c]{@{}c@{}}$\omega: \mathcal{U}(15,50)$, $b': \mathcal{U}(0.1,1)$,\\ $\tau_{m, out}: \mathcal{N}(20,5)$\end{tabular}} & 
\begin{tabular}[c]{@{}c@{}}$\omega: \mathcal{U}(15,35)$, $b': \mathcal{U}(0.1,1)$,\\ $ \alpha: Sigmoid(\mathcal{N}(0,0.1))$\end{tabular}& \multicolumn{2}{c}{\begin{tabular}[c]{@{}c@{}}$\tau_{m}: \mathcal{N}(20,5)$, $\tau_{a}: \mathcal{N}(200,50)$,\\$\tau_{m, out}: \mathcal{N}(20,5)$\end{tabular}}\\
\bottomrule
\label{app:smnist_lc_hyp}
\end{tabularx}

\centering
\begin{tabularx}{\linewidth}{lCCcCC}
\toprule
\multirow{2}{*}{PS-MNIST} & \multicolumn{2}{c}{BRF-RSNN} & BHRF-RSNN& \multicolumn{2}{c}{ALIF-RSNN} \\
& BPTT & TBPTT & BPTT& BPTT & TBPTT \\
\midrule
Network & \multicolumn{2}{c}{1 + 256 (fully recurrent) + 10}  &1 + 256 (fully recurrent) + 10&\multicolumn{2}{c}{1 + 256 (fully recurrent) + 10} \\
Learning rate (Lr) & 0.1 & 0.006 & 0.1& \multicolumn{2}{c}{0.001} \\
Loss function & \multicolumn{2}{c}{NLL} & NLL& \multicolumn{2}{c}{NLL} \\
Minibatch size & \multicolumn{2}{c}{256}&256 & \multicolumn{2}{c}{256}\\
Lr scheduling & \multicolumn{2}{c}{LinearLR}&LinearLR& \multicolumn{2}{c}{LinearLR} \\
Optimizer & \multicolumn{2}{c}{Adam} & RAdam& \multicolumn{2}{c}{Adam} \\
Epochs & \multicolumn{2}{c}{300} & 200 & \multicolumn{2}{c}{300}\\
\begin{tabular}[c]{@{}l@{}}Parameter\\ ~~initialization\end{tabular} & \multicolumn{2}{c}{\begin{tabular}[c]{@{}c@{}}$\omega: \mathcal{U}(15,85)$, $b': \mathcal{U}(0.1,1),$\\ $\tau_{m, out}: \mathcal{N}(20,1)$\end{tabular}} & 
\begin{tabular}[c]{@{}c@{}}$\omega: \mathcal{U}(10,50)$, $b': \mathcal{U}(1,6)$,\\ $\alpha: Sigmoid(\mathcal{N}(0,0.1))$\end{tabular} & \multicolumn{2}{c}{\begin{tabular}[c]{@{}c@{}}$\tau_{m}: \mathcal{N}(20,5)$, $\tau_{a}: \mathcal{N}(200,50)$,\\ $\tau_{m, out}: \mathcal{N}(20,5)$\end{tabular}} \\
\bottomrule
\label{app:psmnist_lc_hyp}
\end{tabularx}

\centering
\begin{tabularx}{\linewidth}{lCCC}
\toprule
ECG-QT & BRF-RSNN & BHRF-RSNN & ALIF-RSNN \\
\midrule
Network & 4 + 36 (fully recurrent) + 6 & 4 + 36 (fully recurrent) + 6& 4 + 36 (fully recurrent) + 6\\
Learning rate (Lr) & 0.1  & 0.3& 0.05 \\
Loss function & NLL & NLL & NLL \\
Minibatch size & 16  & 4& 64\\
Lr scheduling & LinearLR&LinearLR&LinearLR \\
Optimizer & Adam & RAdam& Adam \\
Epochs & 400 & 300 &  400  \\
\begin{tabular}[c]{@{}l@{}}Parameter\\ ~~initialization\end{tabular} &\begin{tabular}[c]{@{}c@{}}$\omega: \mathcal{U}(3,5)$, $b': \mathcal{U}(0.1,1)$,\\ $\tau_{m, out}: \mathcal{N}(20,1)$\end{tabular} &
\begin{tabular}[c]{@{}c@{}}$\omega: \mathcal{U}(7,11)$, $b': \mathcal{U}(0.1,1)$,\\ $\alpha: Sigmoid(\mathcal{N}(0,0.1))$\end{tabular}& \begin{tabular}[c]{@{}c@{}}$\tau_{m}: \mathcal{N}(20,0.5)$, $\tau_{a}: \mathcal{N}(7,0.2)$,\\ $\tau_{m, out}: \mathcal{N}(20,0.5)$\end{tabular} \\
Sub-seq. length & 0 & 0  & 10\\
\bottomrule
\label{app:ecg_lc_hyp}
\end{tabularx}

\centering
\begin{tabularx}{\linewidth}{lCCC}
\toprule
SHD & BRF-RSNN & BHRF-RSNN& ALIF-RSNN \\
\midrule
Network & 700 + 128 (fully recurrent) + 20 & 700 + 128 (fully recurrent) + 20&  700 + 128 (fully recurrent) + 20 \\
Learning rate (Lr) & 0.075   & 0.1& 0.075 \\
Loss function & NLL &NLL& NLL \\
Minibatch size & 32 &32 & 32 \\
Lr scheduling &LinearLR&LinearLR &LinearLR\\
Optimizer & Adam & RMSprop& Adam\\
Epochs & 20 & 20 & 20 \\
\begin{tabular}[c]{@{}l@{}}Parameter\\ ~~initialization\end{tabular} & \begin{tabular}[c]{@{}c@{}}$\omega: \mathcal{U}(5,10)$, $b': \mathcal{U}(2,3)$,\\ $\tau_{m, out}: \mathcal{N}(20,5)$\end{tabular} & \begin{tabular}[c]{@{}c@{}}$\omega: \mathcal{U}(3,10)$, $b': \mathcal{U}(0.1,1)$,\\ $\alpha: Sigmoid(\mathcal{N}(0,0.1))$\end{tabular}&\begin{tabular}[c]{@{}c@{}}$\tau_{m}: \mathcal{N}(20,5)$, $\tau_{a}: \mathcal{N}(150,10)$,\\ $\tau_{m, out}: \mathcal{N}(20,5)$\end{tabular} \\
Sub-seq. length & 0& 0 & 10  \\
\bottomrule
\end{tabularx}
\end{table}
Due to the BHRF-RSNN being in the preliminary phase of its exploration, we initially used the sigmoid function for the leaky integrator decay constant $\alpha$, as it ensured the decay to be in the range of 0 and 1.

\subsection{Learning speed-up with forward gradient injection (FGI) and TorchScript} \label{sec:fgi}

As a further exploration, we apply forward gradient injection (FGI) from \citet{otte2024fgi} for modeling the surrogate gradient functions. Let $x$ be a computation node, $f$ a non-differentiable function (e.g. step function), $g'$ a function that we want to use as derivative for $f$, and $\operatorname{sg}$ the stop gradient operator. Instead of overriding the \texttt{backward()} method, we can just write:
\vspace{-0.1cm}
\begin{align}
    h &= x \cdot \operatorname{sg}({g'(x)})\\
    y &= h - \operatorname{sg}(h) + \operatorname{sg}(f(x))
\end{align}
\vspace{-0.1cm}
With FGI + TorchScript we achieve speed-ups by a factor of more than two for all dataset as shown in \autoref{tab:fgi}, while maintaining performance.

\begin{table}[h!]
\small
\caption{BRF-RSNN training speed up with model optimization methods over standard backward() baseline.}
\label{tab:fgi}
\centering    
\begin{tabularx}{\linewidth}{cccCcCcCcC}
\toprule
\multirow{2}{*}{Optimization} & \multirow{2}{*}{Gradient}&\multicolumn{2}{c}{S-MNIST} & \multicolumn{2}{c}{PS-MNIST}&\multicolumn{2}{c}{ECG} & \multicolumn{2}{c}{SHD}\\
& & Time (hrs) & Ratio & Time (hrs) &  Ratio & Time (hrs) & Ratio & Time (min) &  Ratio\\
\midrule
- & backward() & 23.8 & $1\times$& 31.0 & $1\times$& 8.5 & $1\times$& 40.9 & $1\times$\\
TorchScript& backward() & 11.3 & $2.1\times$& 13.8 &$2.2\times$ & 4.1 & $2.1\times$& 16.4 & $2.5\times$\\
TorchScript  & FGI & 10.2 &$\mathbf{2.3}\times$ & 12.8 & $\mathbf{2.4}\times$& 3.5 & $\mathbf{2.4}\times$ & 14.0 & $\mathbf{2.9}\times$ \\
\bottomrule
\end{tabularx}
    \label{tabs:script_fgi}
\end{table}

\end{document}